\documentclass[lettersize,journal]{IEEEtran}
\usepackage{amsmath,amsfonts}
\usepackage{algorithmic}
\usepackage{algorithm}
\usepackage{array}
\usepackage[caption=false,font=normalsize,labelfont=sf,textfont=sf]{subfig}
\usepackage{textcomp}
\usepackage{stfloats}
\usepackage{url}
\usepackage{bm}
\usepackage{color}
\usepackage{multirow}
\usepackage{booktabs}
\usepackage{verbatim}
\usepackage{graphicx}
\usepackage{cite}
\usepackage{hyperref,MnSymbol}
\begin{document}

\title{PKU-AIGIQA-4K: A Perceptual Quality Assessment Database for Both Text-to-Image \\and Image-to-Image AI-Generated Images}

\author{Jiquan Yuan, Fanyi Yang, Jihe Li, Xinyan Cao, Jinming Che, Jinlong Lin, Xixin Cao
        % <-this % stops a space
\thanks{This work was supported by PKU-HW under Grant 8410101869. Corresponding author: Xixin Cao}
\thanks{Jiquan Yuan, Fanyi Yang, Jihe Li, Xinyan Cao, Jinming Che, Jinlong Lin, and Xixin Cao are with the School of Software \& Microelectronics, Peking University, Beijing 102600, China (email: {yuanjiquan, yangfanyi,  lijh, caoxinyan, chejinming}@stu.pku.edu.cn, {linjl,cxx}@ss.pku.edu.cn)}% <-this % stops a space
}

\markboth{Journal of \LaTeX\ Class Files,~Vol.~, No.~, April~2024}%
{Shell \MakeLowercase{\textit{et al.}}: A Sample Article Using IEEEtran.cls for IEEE Journals}


\maketitle

\begin{abstract}
In recent years, image generation technology has rapidly advanced, resulting in the creation of a vast array of AI-generated images (AIGIs). However, the quality of these AIGIs is highly inconsistent, with low-quality AIGIs severely impairing the visual experience of users. Due to the widespread application of AIGIs, the AI-generated image quality assessment (AIGIQA), aimed at evaluating the quality of AIGIs from the perspective of human perception, has garnered increasing interest among scholars. Nonetheless, current research has not yet fully explored this field. We have observed that existing databases are limited to images generated from single scenario settings. Databases such as AGIQA-1K, AGIQA-3K, and AIGCIQA2023, for example, only include images generated by text-to-image generative models. This oversight highlights a critical gap in the current research landscape, underscoring the need for dedicated databases catering to image-to-image scenarios, as well as more comprehensive databases that encompass a broader range of AI-generated image scenarios. Addressing these issues, we have established a large scale perceptual quality assessment database for both text-to-image and image-to-image AIGIs, named PKU-AIGIQA-4K. We then conduct a well-organized subjective experiment to collect quality labels for AIGIs and perform a comprehensive analysis of the PKU-AIGIQA-4K database. Regarding the use of image prompts during the training process, we propose three image quality assessment (IQA) methods based on pre-trained models that include a no-reference method NR-AIGCIQA, a full-reference method FR-AIGCIQA, and a partial-reference method PR-AIGCIQA. Finally, leveraging the PKU-AIGIQA-4K database, we conduct extensive benchmark experiments and compare the performance of the proposed methods and the current IQA methods. The PKU-AIGIQA-4K database and code will be released on https://github.com/jiquan123/AIGIQA4K.
\end{abstract}


\section{Introduction}
\IEEEPARstart{I}{n} the current era marked by rapid advancements in computer vision and the internet, images have become the primary medium for information acquisition and dissemination. With the continuous development of image generation technology, a plethora of sophisticated image generative models have emerged in the market, such as Midjourney~\cite{r1}, Stable Diffusion~\cite{r2}, Glide~\cite{nichol2021glide},
Lafite~\cite{zhou2022lafite}, DALLE~\cite{dalle}, Unidiffuser~\cite{bao2023unidiff}, Controlnet~\cite{zhang2023addingcontrolnet}, \textit{etc.} These models enable the convenient creation of a vast quantity of AI-generated images (AIGIs). However, AIGIs may exhibit unique distortions not found in natural images, such as unrealistic structures, irregular textures and shapes, and AI artifacts~\cite{zhang2023perceptual,AGIQA-3K,wang2023aigciqa2023}, \textit{etc.}, which can degrade image quality and, consequently, impair the visual experience of users. Moreover, not all AIGIs meet the requirements of the real world, often necessitating processing, adjustment, refinement, or filtering before their application in practical scenarios. Thus, developing efficient algorithms to assess the quality of AIGIs is crucial. On one hand, this can facilitate the quality screening of a large volume of AIGIs to meet the demand for high-quality images. On the other hand, it can serve to evaluate existing image generative models, promoting the creation of higher quality images.

Due to the wide application of AIGIs, the AI-generated image quality assessment (AIGIQA), aimed at evaluating the quality of AIGIs from the perspective of human perception, has garnered increasing interest among scholars. Over the past few years, numerous dedicated AIGIQA databases and algorithms~\cite{zhang2023perceptual,AGIQA-3K,wang2023aigciqa2023,yuan2023pscr,yuan2024tier,yang2024aigcoiqa2024,Li2024AIGIQA20KAL} have been proposed to foster development in this field. However, existing research has not yet fully explored this domain. It has been observed that existing databases focus solely on images generated from single scenario settings. Databases such as AGIQA-1K~\cite{zhang2023perceptual}, AGIQA-3K~\cite{AGIQA-3K}, and AIGCIQA2023~\cite{wang2023aigciqa2023}, AIGCOIQA2024\cite{yang2024aigcoiqa2024}, AIGIQA-20K\cite{Li2024AIGIQA20KAL}, \textit{etc.}, for instance, only collect AIGIs produced by text-to-image generative models. This oversight highlights a critical gap in the current research landscape, underscoring the need for dedicated databases catering to image-to-image scenarios, as well as more comprehensive databases that encompass a broader range of AI-generated image scenarios. The establishment of such databases is imperative to enable a more holistic assessment for AI-generated image quality.

\begin{figure*}[!t]
\centering
\includegraphics[width=7in]{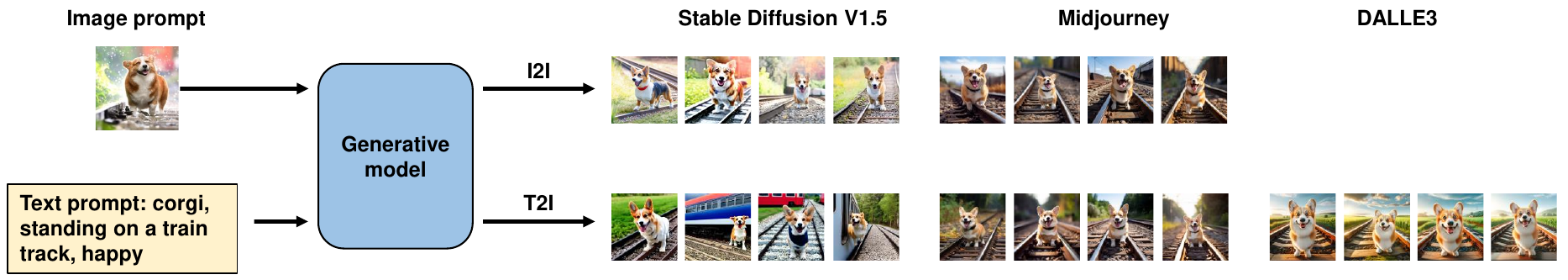}
\caption{Illustration of the text-to-image and image-to-image generation in the PKU-AIGIQA-4K database.}
\label{fig_1}
\end{figure*}

To address the above issues, we first establish a large scale perceptual quality assessment database that includes both text-to-image and image-to-image AIGIs, named PKU-AIGIQA-4K. To our best knowledge, this is the first perceptual quality assessment database that encompasses both text-to-image and image-to-image AIGIs. Specifically, we utilize the three most popular image generative models Midjourney~\cite{r1}, Stable Diffusion V1.5~\cite{r2}, and DALLE3~\cite{DALLE3} for image generation. Stable Diffusion V1.5 and Midjourney serve as both text-to-image and image-to-image generative models, while DALLE3 function solely as image-to-image generative model. For each prompt, we randomly generate four images per model. The PKU-AIGIQA-4K database we constructed comprises two subsets, I2IQA and T2IQA, totaling 4,000 images. The I2IQA contains 1,600 images (\text{4 images} $\times$ \text{2 models} $\times$ \text{200 prompts}) generated by image-to-image generative models, and the T2IQA consists of 2,400 images (\text{4 images} $\times$ \text{3 models} $\times$ \text{200 prompts}) produced by text-to-image generative models. Fig.\ref{fig_1} demonstrates the text-to-image and image-to-image generation of the PKU-AIGIQA-4K database. We then conduct a well-organized subjective experiment to collect quality labels for the AIGIs and carry out a comprehensive analysis of the PKU-AIGIQA-4K database. TABLE \ref{table1} compares the PKU-AIGIQA-4K database with existing AIGIQA databases.

\begin{table}[t]
\caption{Comparison with several existing AIGIQA databases.}
\centering
\begin{tabular}{lccc}
\toprule
Database & Models & Category & AIGIs \\ \hline
AGIQA-1K\cite{zhang2023perceptual}       & 2  &T2I AIGIs only                           & 1080          \\ 
AGIQA-3K\cite{AGIQA-3K}       & 6     &T2I AIGIs only                           & 2982          \\ 
AIGCIQA2023\cite{wang2023aigciqa2023}   & 6    &T2I AIGIs only                           & 2400          \\ 
AIGCOIQA2024\cite{yang2024aigcoiqa2024}       & 5     &T2I AIGIs only                           & 300          \\ 
AIGIQA-20K\cite{Li2024AIGIQA20KAL}   & 15    &T2I AIGIs only                           & 20000          \\
\textbf{PKU-AIGIQA-4K}             & 3         &both T2I and I2I AIGIs                   & 4000          \\ 
\bottomrule
\end{tabular}
\label{table1}
\end{table}

Moreover, due to the inclusion of both text-to-image and image-to-image AIGIs in our constructed PKU-AIGIQA-4K database, existing image quality assessment (IQA) methods often fail to fully utilize the information present in the image prompts within the database. Specifically, the lack of reference images for text-to-image AIGIs renders full-reference IQA methods inapplicable on the PKU-AIGIQA-4K database; meanwhile, employing no-reference IQA methods does not adequately exploit the information from the image prompts corresponding to the image-to-image AIGIs. To fully leverage the available image information and simultaneously assess both text-to-image and image-to-image AIGIs, we propose a partial-reference IQA method, named PR-AIGCIQA, which can be employed when only partial AIGIs have reference counterparts. Specifically, we adopt a solution from natural language processing (NLP) for handling variable-length text inputs by padding text-to-image AIGIs with a zero vector of identical size to ensure uniform input vector dimensions. In subsequent calculations, we use a mask vector to ignore this padded portion, thus disregarding this irrelevant information. When evaluating AIGIs generated from single scenario settings in the PKU-AIGIQA-4K database's subsets I2IQA and T2IQA, depending on whether image prompts are used as reference images during training and testing, we design a no-reference IQA method NR-AIGCIQA and a full-reference IQA method FR-AIGCIQA based on pre-trained models. Finally, leveraging the PKU-AIGIQA-4K database, we conduct extensive benchmark experiments and compare the performance of the proposed methods and the current IQA methods.

The main contributions of this paper are summarized as follows:

\begin{itemize}

\item We establish the first perceptual quality assessment database PKU-AIGIQA-4K that includes both text-to-image and image-to-image AIGIs for AIGIQA task, with its subset I2IQA being the first dedicated database for assessing the quality of image-to-image AIGIs. It comprises two subsets, I2IQA and T2IQA, totaling 4,000 images. 

\item We conduct a well-organized subjective experiment to collect quality labels for AIGIs and carry out a comprehensive analysis of the PKU-AIGIQA-4K database.

\item We propose no-reference, full-reference, and partial-reference IQA methods based on pre-trained models, named NR-AIGCIQA, FR-AIGCIQA, and PR-AIGCIQA, respectively.

\item Extensive benchmark experiments are performed on the PKU-AIGIQA-4K database to compare the performance of the proposed methods and the current IQA methods.

\end{itemize}

This paper is an extension of our previous work\cite{yuan2023pku}, the major extensions include: 1) the addition of 2,400 text-to-image AIGIs to the PKU-I2IQA\cite{yuan2023pku} database; 2) the introduction of a partial-reference IQA method based on pre-trained models, named PR-AIGCIQA; 3) extensive utilization of the text prompts and their corresponding AIGIs, combining the TIER\cite{yuan2024tier} method with the methods presented in this paper for experiments. 

\begin{figure*}[!t]
\centering
\includegraphics[width=7in]{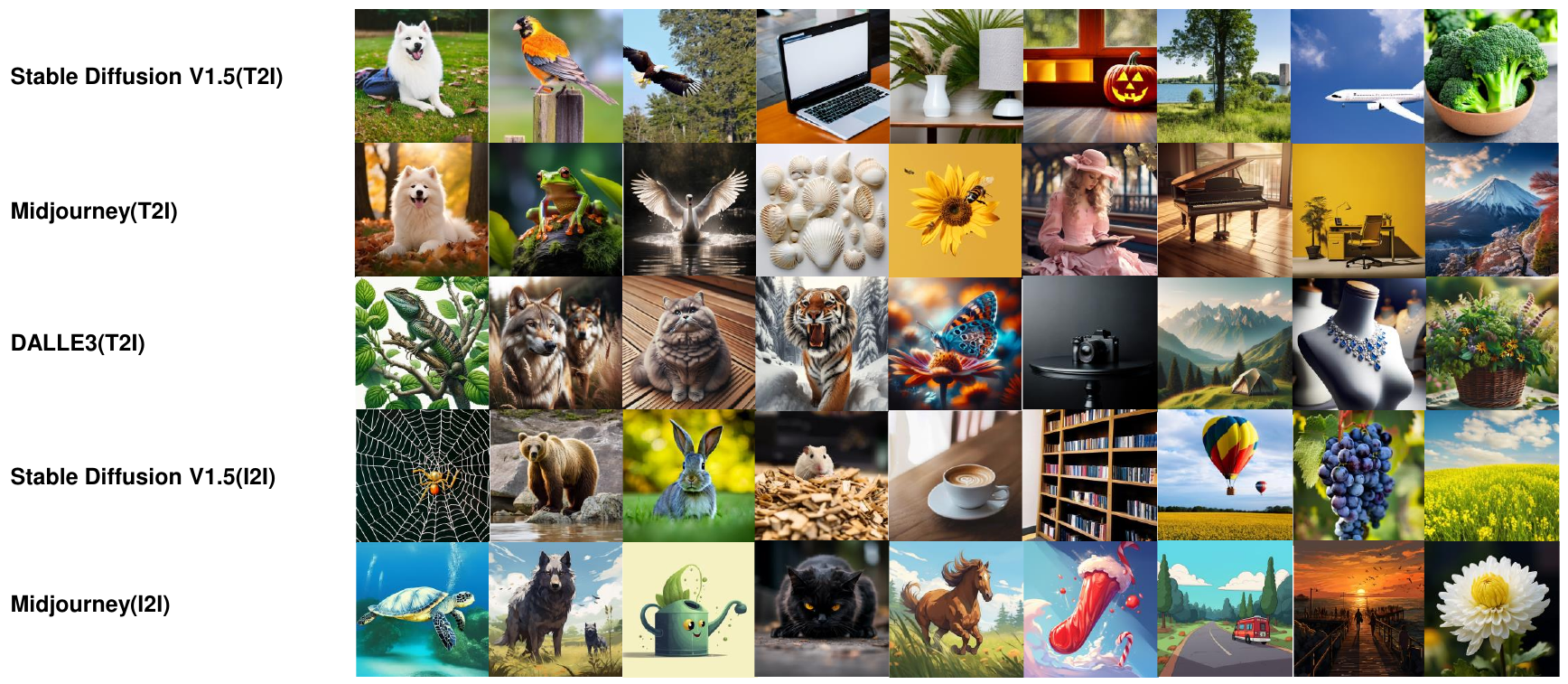}
\caption{Various scenes and styles of AIGIs sampled from the PKU-AIGIQA-4K database generated by Midjourney, Stable Diffusion V1.5, and DALLE3, including text-to-image and image-to-image AIGIs.}
\label{Sample}
\end{figure*}

\section{Related Work}
\subsection{Image Quality Assessment.}In the past few years, researchers have proposed numerous image quality assessment (IQA) methods~\cite{FGIQA,nr-fr,lao2022attentions,fr3,chen2021full,cheon2021IQT,nr-iqa,liu2017rankiqa,MetaIQA,yang2022maniqa,cnriqa,bmsb,mdaqa,DBCNN,hyperiqa,li2020LinearityIQA,VGCN}. The IQA methods can be categorized into FR-IQA methods~\cite{nr-fr,lao2022attentions,fr3,chen2021full,cheon2021IQT} and NR-IQA methods~\cite{nr-iqa,liu2017rankiqa,MetaIQA,yang2022maniqa,cnriqa,bmsb,mdaqa,DBCNN,hyperiqa,li2020LinearityIQA,VGCN}, depending on whether a reference image is used during the prediction process. Full-reference methods often achieve higher prediction accuracy compared to no-reference methods, as the inclusion of a reference image allows the computer to extract more effective features during the prediction process. Many classical IQA models initially employ methods based on manually extracted features~\cite{nr-iqa,r19,r20}. However, with the rapid development of convolutional neural networks, methods based on deep learning for feature extraction~\cite{nr-fr,lao2022attentions,fr3,chen2021full,cheon2021IQT,liu2017rankiqa,MetaIQA,yang2022maniqa,cnriqa,bmsb,mdaqa, DBCNN,hyperiqa,li2020LinearityIQA,VGCN} have led to significant performance improvements. As a branch of IQA, the AI-generated image quality assessment (AIGIQA) task has garnered increasing interest among scholars. Nonetheless, current research has not yet fully explored this field. Previously, AIGIQA relies on automatic measures like Inception Score (IS)~\cite{r6}, Fréchet Inception Distance (FID)~\cite{r7}, and CLIP Score~\cite{hessel2021clipscore}, \textit{etc.} 
Gu \textit{et al.}~\cite{gu2020giqa} pioneer the research direction of assessing the quality of AIGIs, aiming to develop algorithms for automatically scoring the quality of AIGIs.
Recently, Mayu Otani \textit{et al.}~\cite{r8} from the Japanese internet giant Cyber Agent conduct a detailed investigation and experiments on evaluation metrics for AIGIQA. They find that current evaluation metrics are limited to express human perception, especially in terms of FID~\cite{r7} and CLIP Score~\cite{hessel2021clipscore}, and are unable to evaluate the state-of-the-art image generative models. Zhang \textit{et al.}~\cite{zhang2023perceptual} establish the first human perception-based text-to-image database for AIGIQA, named AGIQA-1K. It consists of 1,080 AIGIs generated by 2 diffusion models~\cite{r2}. Through well-organized subjective experiments, human subjective perception evaluations of AIGIs are introduced to collect quality labels for AIGIs. They further conduct benchmark experiments to evaluate the performance of the current IQA models~\cite{he2016resnet,bmsb,mdaqa}. Li \textit{et al.}~\cite{AGIQA-3K} consider six representative generative models and build the most comprehensive AIGI subjective quality database AGIQA-3K. This is the first database that covers AIGIs from GAN/auto regression/diffusion-based model altogether. Wang \textit{et al.}~\cite{wang2023aigciqa2023} establish a large-scale AIGCIQA database, named AIGCIQA2023. They utilize 100 prompts and generate over 2000 images based on six state-of-the-art text-to-image generative models~\cite{bao2023unidiff,dalle,zhou2022lafite,r2,nichol2021glide,zhang2023addingcontrolnet}. A well-organized subjective experiment is conducted on these images to evaluate human preferences for each image from the perspectives of quality, authenticity, and text-image correspondence. Finally, they perform benchmark experiments on this large-scale database to evaluate the performance of several state-of-the-art IQA models~\cite{cnriqa,simonyan2014vgg,he2016resnet,nr-fr}. Moreover, Yang \textit{et al.}~\cite{yang2024aigcoiqa2024} establish a large-scale 
database named AIGCOIQA2024 for AI-generated omnidirectional images and construct a comprehensive benchmark. Li \textit{et al.}~\cite{Li2024AIGIQA20KAL} create the largest fine-grained
AIGI subjective quality database to date with 20,000 AIGIs
and 420,000 subjective ratings, known as AIGIQA-20K.  Although these efforts have advanced the development of AIGIQA, there remain issues to be addressed, such as how to encompass AIGIs across various scenarios as comprehensively as possible and the establishment of more dedicated databases. In this paper, we first establish a large scale perceptual quality assessment database named PKU-AIGIQA-4K, which contains both text-to-image AIGIs and image-to-image AIGIs. Based on three scenarios during the training process: 1) no AIGIs use image prompts as references; 2) all images use image prompts as references; 3) partial images use image prompts as references,  we propose three image quality assessment (IQA) methods based on pre-trained models that include a no-reference IQA method NR-AIGCIQA, a full-reference IQA method FR-AIGCIQA, and a partial-reference IQA method PR-AIGCIQA.

\subsection{Visual Backbone.}Visual backbone networks are fundamental and crucial components in computer vision, employed for feature extraction and representation in image processing tasks. These network models typically consist of multiple layers and modules designed to extract and represent features from input images, supporting various computer vision tasks such as object detection, image classification, semantic segmentation, \textit{etc.} In the last decade, deep learning has seen remarkable progress, especially after the introduction of ImageNet~\cite{russakovsky2015imagenet} by Fei-Fei Li and her colleagues at Stanford University. This has significantly advanced deep learning's role in various computer vision tasks. We've seen the development of multiple visual backbone models, such as CNN-based ones like VGG~\cite{simonyan2014vgg}, GoogleNet~\cite{szegedy2015google}, ResNet~\cite{he2016resnet}, and transformer-based ones like ViT~\cite{dosovitskiy2020vit}, Swin Transformer~\cite{liu2021swin}, \textit{etc.} In this paper, we employ several backbone network models pre-trained on the ImageNet~\cite{russakovsky2015imagenet} as feature extraction networks.

\section{Database Construction and Analysis}
\subsection{AIGI Collection}

To ensure the diversity of the generated content, we select 200 categories from the famous large-scale image database ImageNet~\cite{russakovsky2015imagenet} in the field of computer vision. Subsequently, we collect corresponding images from the high-resolution image website Pixabay~\cite{Pixabay} based on the selected categories to serve as image prompts for image-to-image generative models. \textbf{It is explicitly stated that we use the royalty-free images from this website}. These prompts include images of various scenes such as animals, plants, furniture, and natural landscapes, \textit{etc.} We then use CLIP~\cite{radford2021clip} to perform reverse deduction to obtain text prompts from image prompts, which are used for both text-to-image and image-to-image generation.

We employ three most popular image generative models Midjourney~\cite{r1}, Stable Diffusion V1.5~\cite{r2} and DALLE3~\cite{DALLE3} as our AIGI generative models.  
Midjourney and Stable Diffusion V1.5 serve both as text-to-image and image-to-image generative models, whereas DALLE-3 is utilized solely as text-to-image generative model. For each prompt, we generate four images randomly for each generative model. The PKU-AIGIQA-4K database we constructed comprises two subsets, I2IQA and T2IQA, totaling 4,000 images. The I2IQA contains 1,600 images (\text{4 images} $\times$ \text{2 models} $\times$ \text{200 prompts}) generated by image-to-image generative models, and the T2IQA consists of 2,400 images (\text{4 images} $\times$ \text{3 models} $\times$ \text{200 prompts}) produced by text-to-image generative models. Various scenes and styles of images sampled from the AIGIQA-4K database are shown in Fig.\ref{Sample}.

\subsection{Subjective Experiment}
To evaluate the image quality of the PKU-AIGIQA-4K database and obtain Mean Opinion Scores (MOSs), subjective experiments are conducted following the guidance of ITU-R BT.500-14~\cite{ITUR}. Following previous work~\cite{wang2023aigciqa2023}, evaluators are asked to express their preferences for the displayed AIGIs from three aspects: quality, authenticity, and text-image correspondence. Quality score is assessed based on clarity, color, brightness, and contrast of AIGIs, along with sharpness of contours, detail richness, and overall aesthetic appeal, \textit{etc.} Authenticity score focuses on whether the AIGIs look real and  whether evaluators could distinguish that the images are generated by AI-based generative models or not. Text-image correspondence scores refers to the matching degree between the AIGIs and the corresponding text prompts.

\begin{figure}[!h]
\centering
\includegraphics[width=3in]{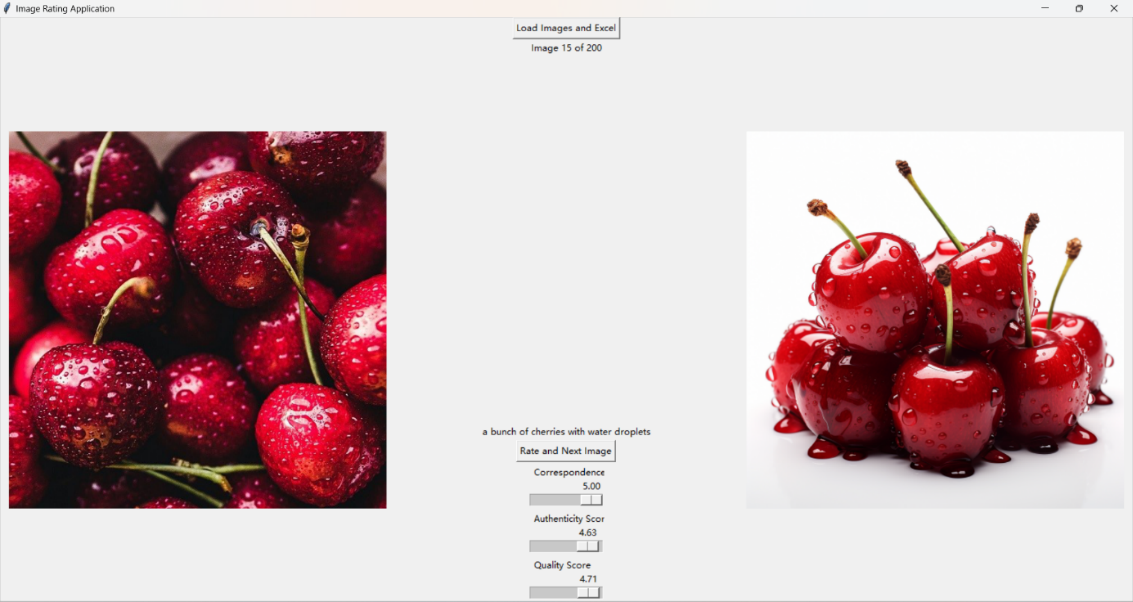}
\caption{An example of the subjective evaluation interface. Evaluators can evaluate the quality of AIGIs by comparing the reference image on the left with the to-be-evaluated AIGIs on the right. They can use the sliders below to record the text-image correspondence score, authenticity score, and quality score.}
\label{Python intreface}
\end{figure}

We employ a Python Tkinter-based graphical interface to display AIGIs in their native $512\times512$ resolution on the computer screen in a random sequence, as illustrated in Fig.\ref{Python intreface}. Using this interface, evaluators rate AIGIs on a 0 to 5 scale with 0.01 increments. Unlike prior studies~\cite{zhang2023perceptual,wang2023aigciqa2023,AGIQA-3K}, we integrate image prompts as reference images into the graphical interface. This enables evaluators to conduct more accurate evaluation by directly comparing these images with the AIGIs under review. 

Twenty graduate students participate in our experiment, which is divided into 20 stages to keep each evaluation session around an hour. In each stage, evaluators need to evaluate 200 AIGIs.

\begin{figure}[!t]
\centering
\subfloat[Text prompt: a bunch of cherries with water droplets]{\includegraphics[width=3in]{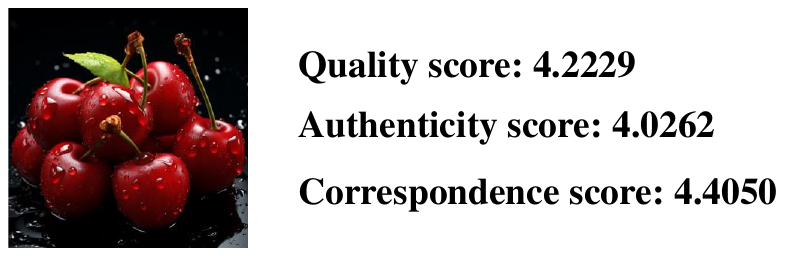}%
\label{cherries}}
\hfil
\subfloat[Text prompt: jellyfishs floating in the water with a black background]{\includegraphics[width=3in]{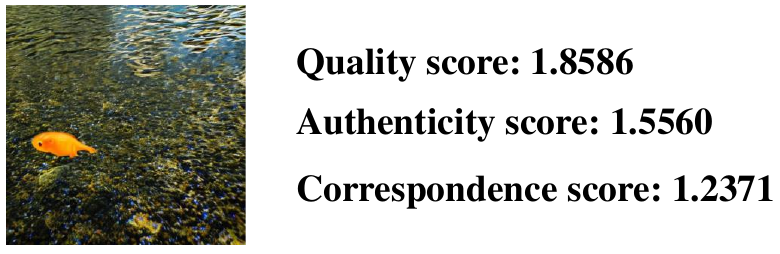}%
\label{jellyfishs}}
\caption{Illustration of AIGIs from three evaluation perspectives. (a) Top
AIGI has better quality, authenticity and correspondence. (b) Bottom AIGI has worse quality, authenticity and correspondence}
\label{example}
\end{figure}

\begin{figure*}[!t]
\centering
\subfloat[]{\includegraphics[width=2in]{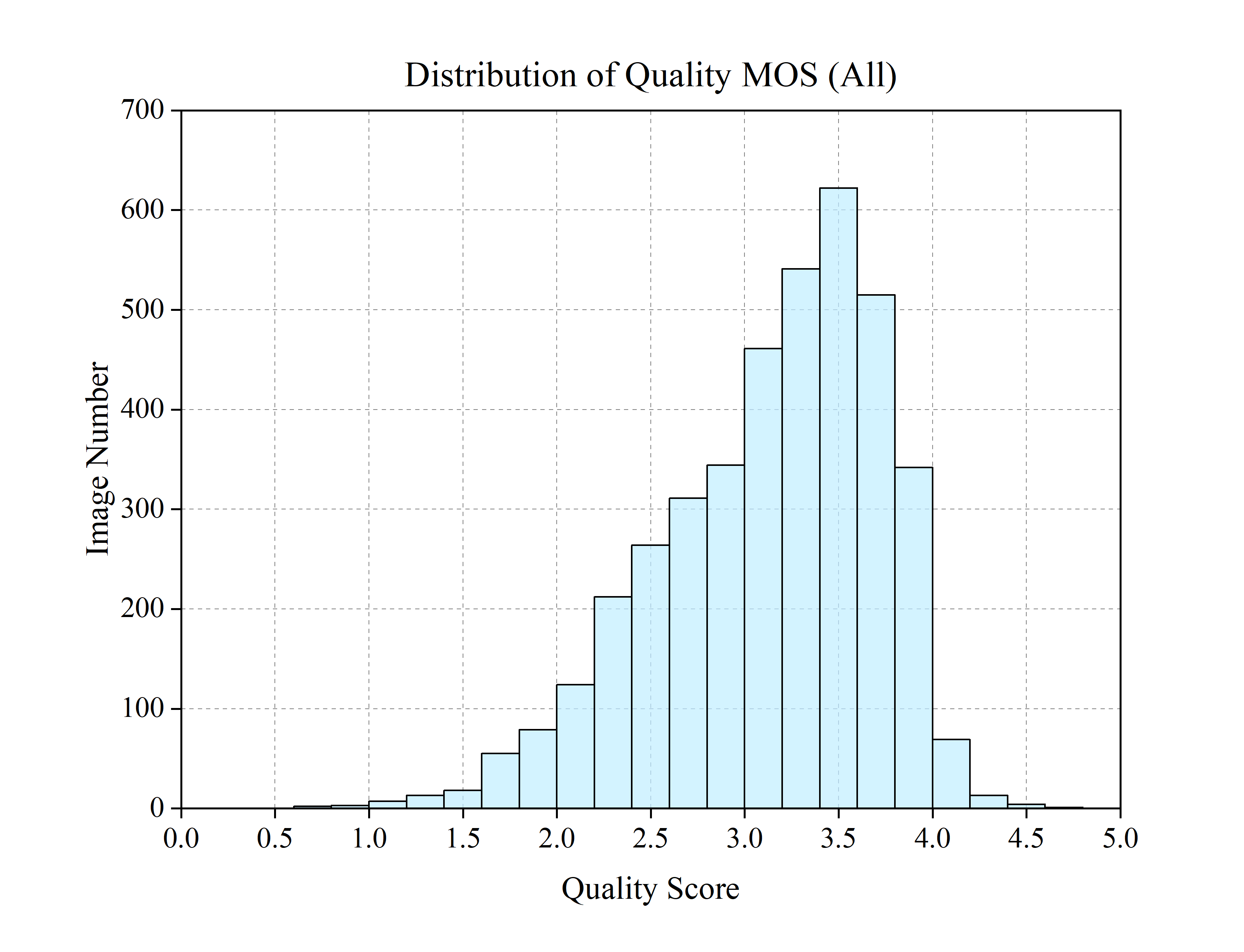}%
\label{M1}}
\hfil
\subfloat[]{\includegraphics[width=2in]{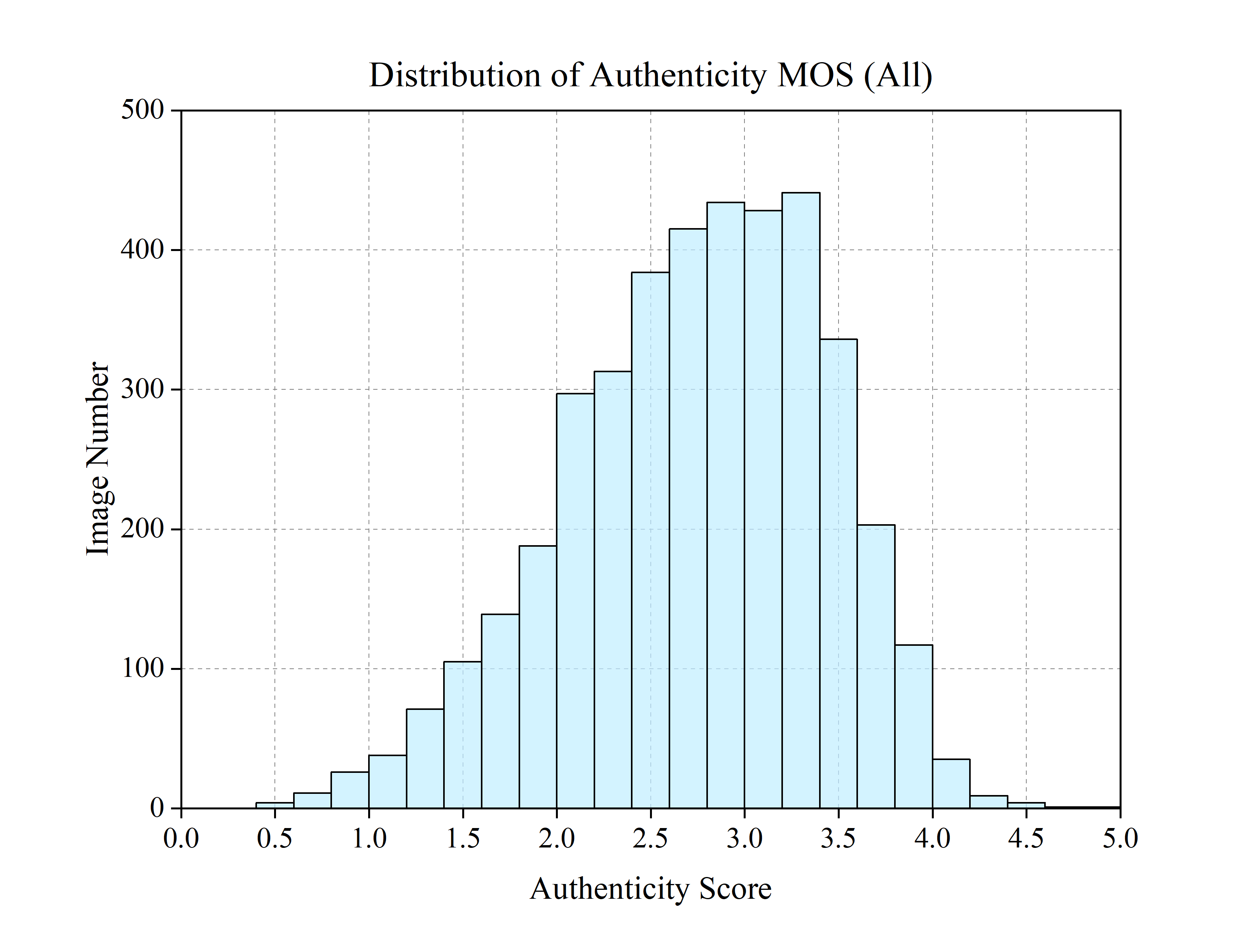}%
\label{M2}}
\hfil
\subfloat[]{\includegraphics[width=2in]{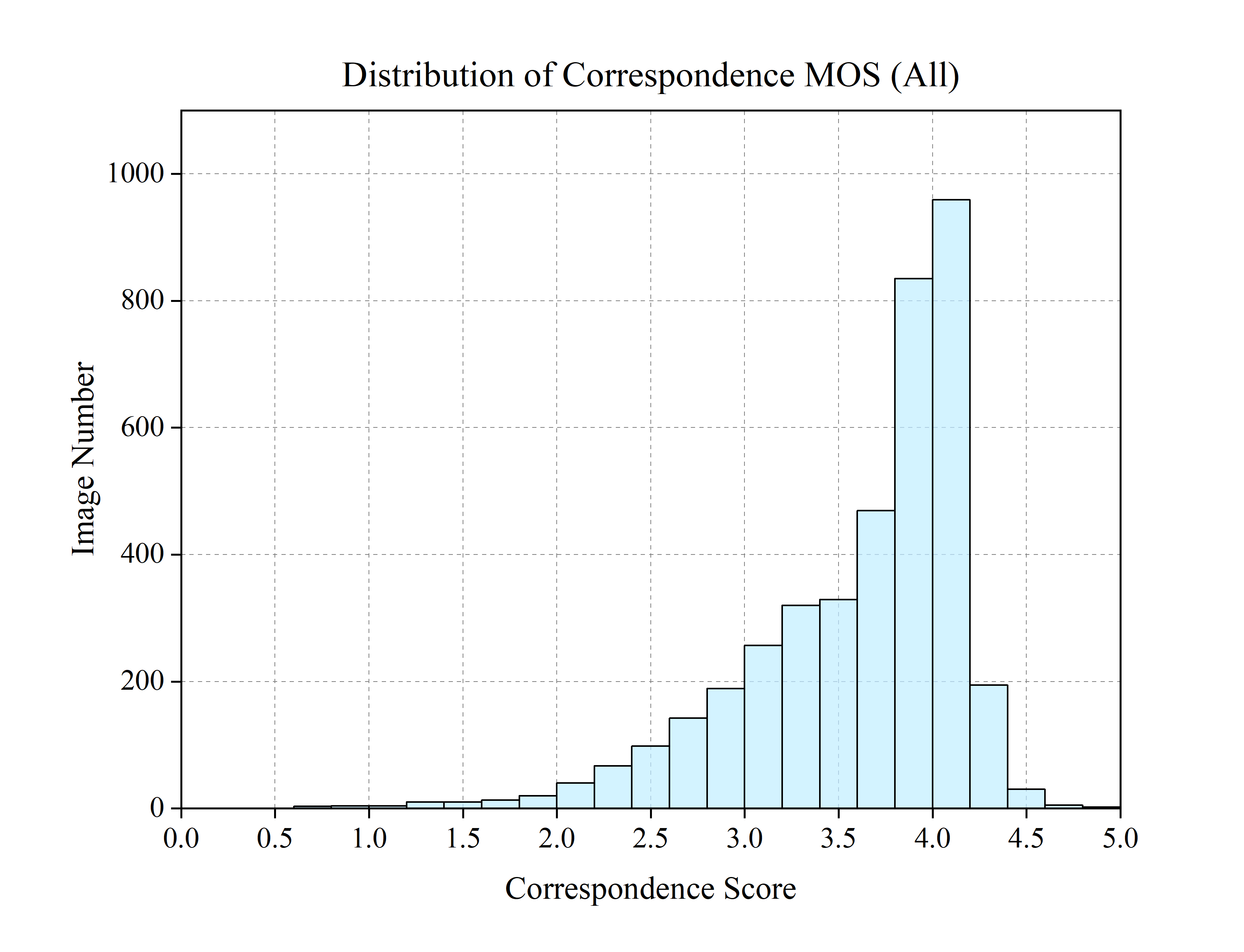}%
\label{M3}}
\hfil
\subfloat[]{\includegraphics[width=2in]{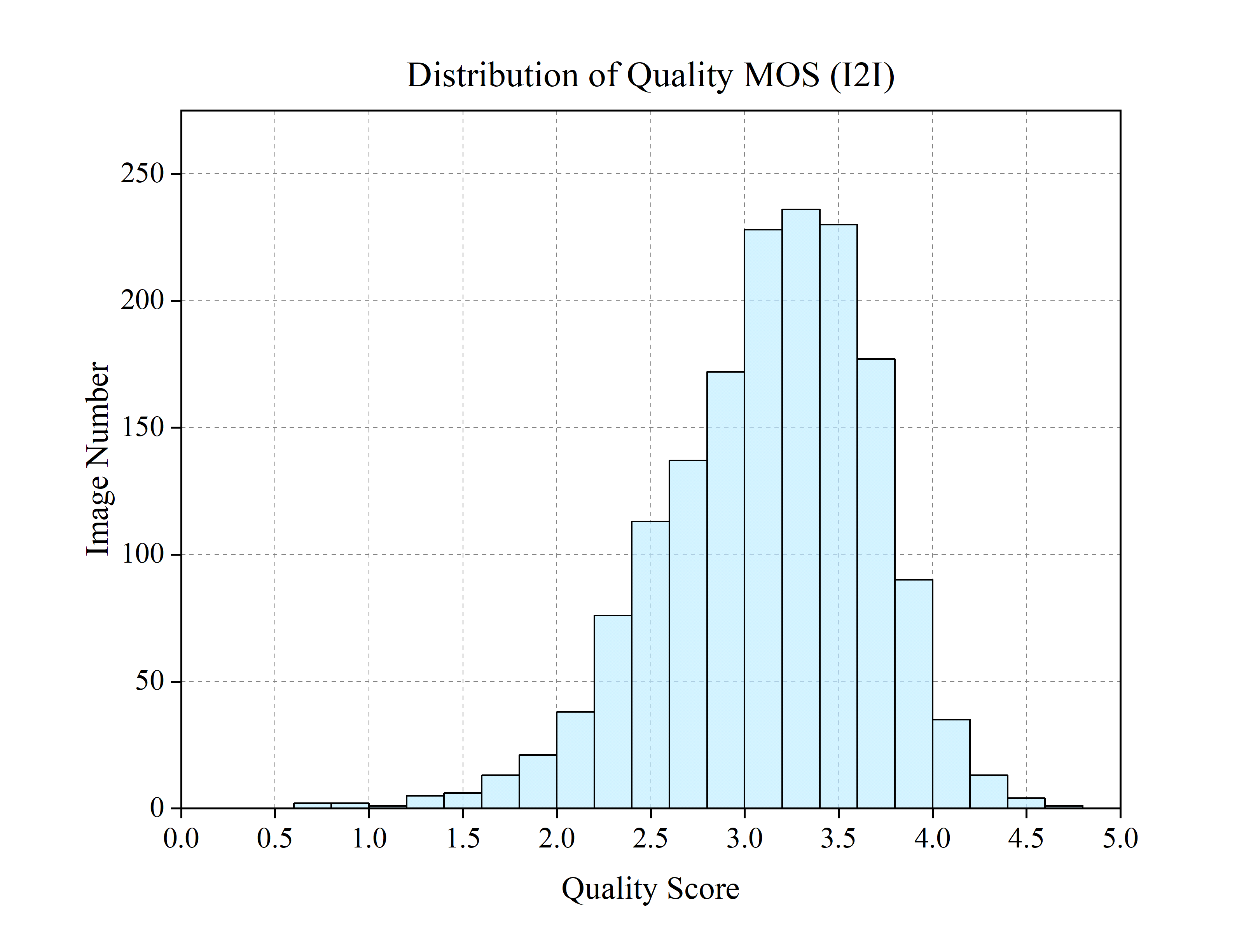}%
\label{M4}}
\hfil
\subfloat[]{\includegraphics[width=2in]{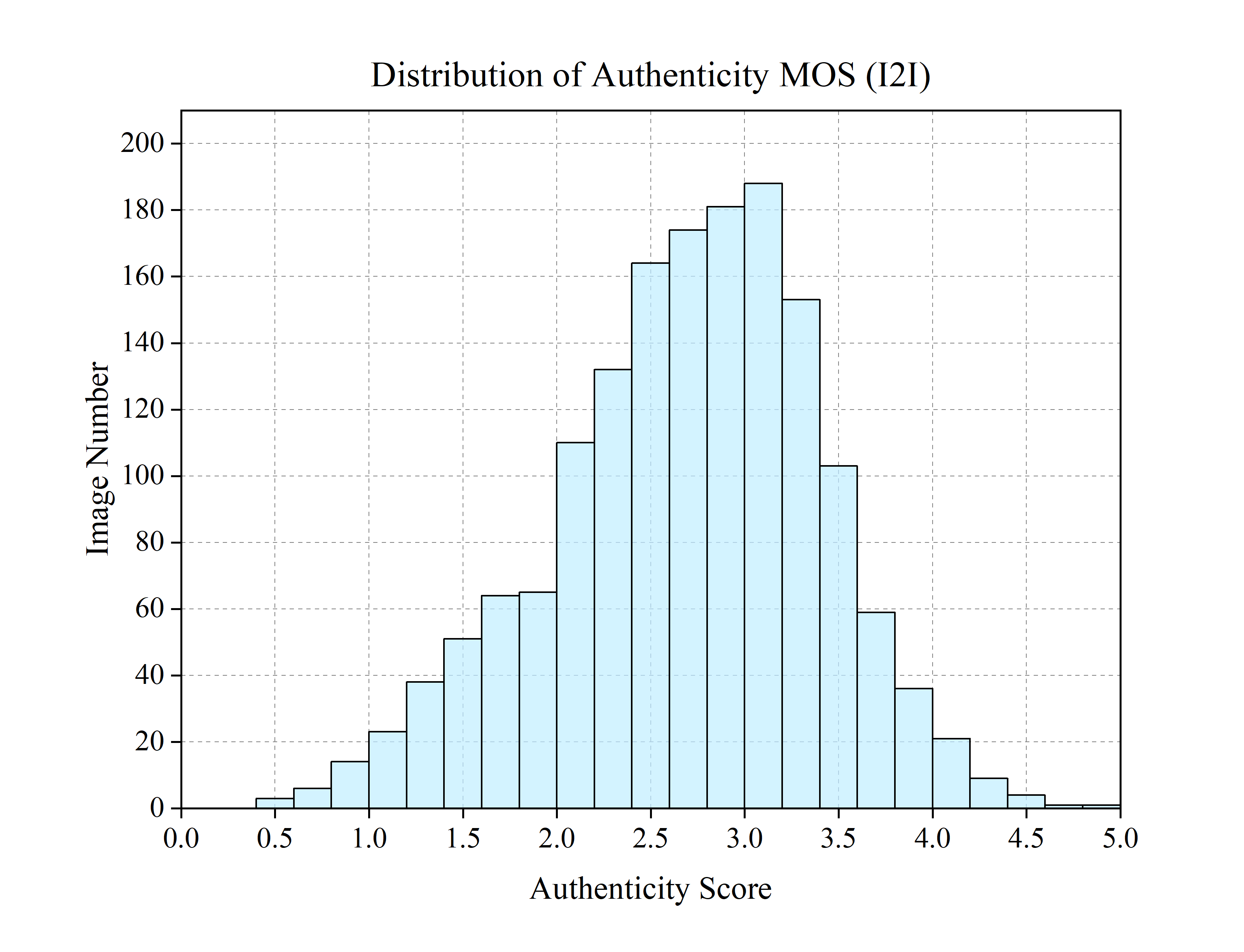}%
\label{M5}}
\hfil
\subfloat[]{\includegraphics[width=2in]{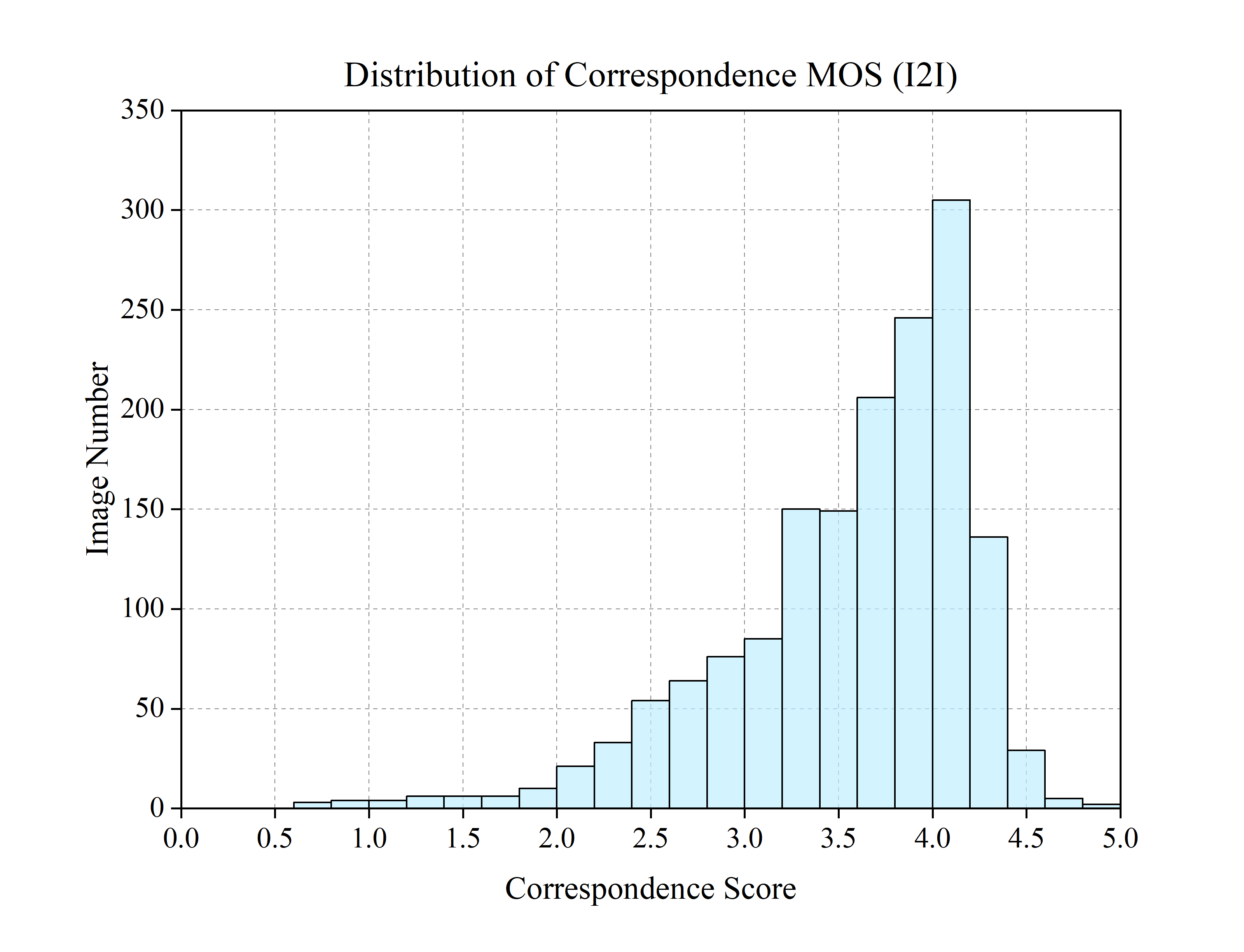}%
\label{M6}}
\hfil
\subfloat[]{\includegraphics[width=2in]{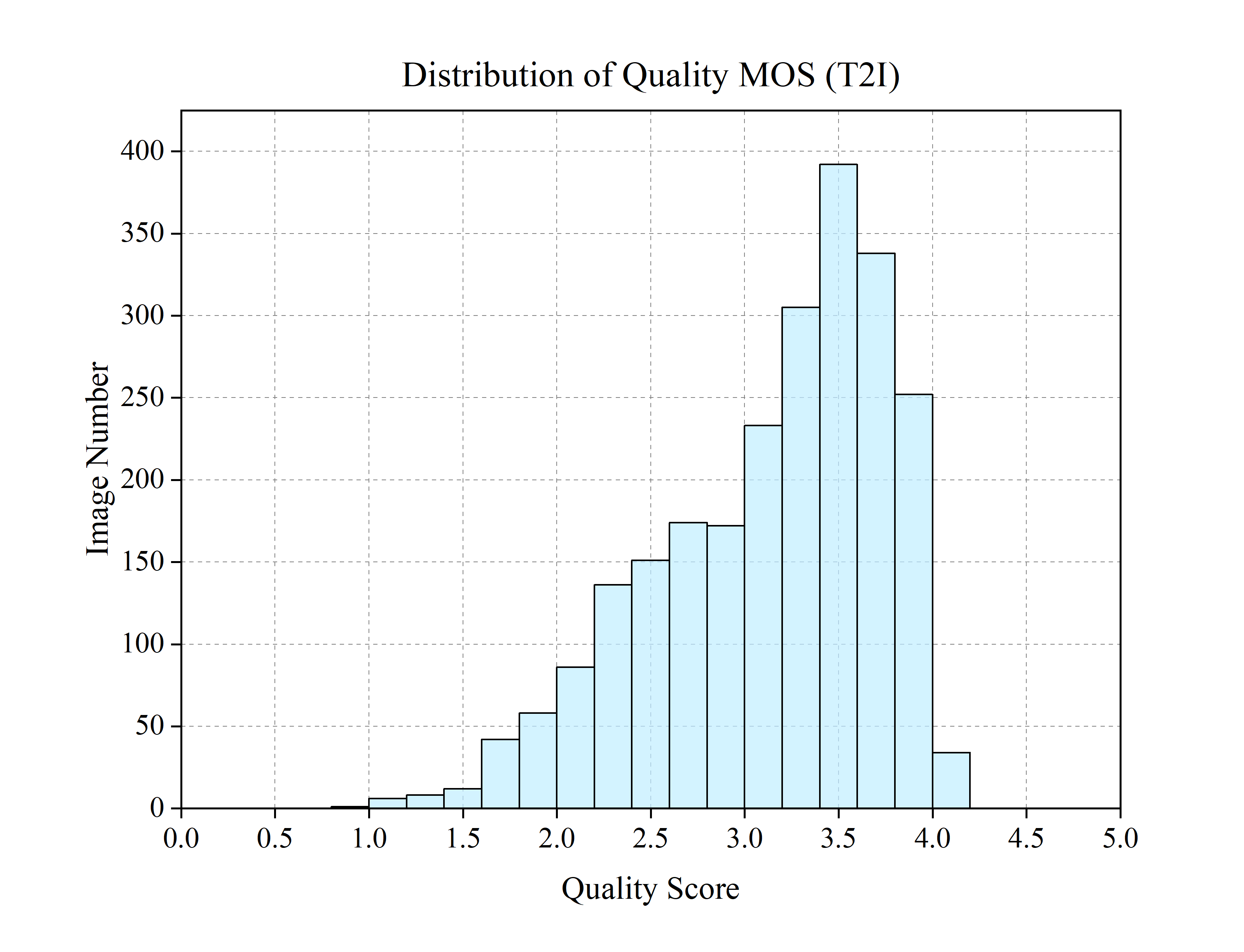}%
\label{M7}}
\hfil
\subfloat[]{\includegraphics[width=2in]{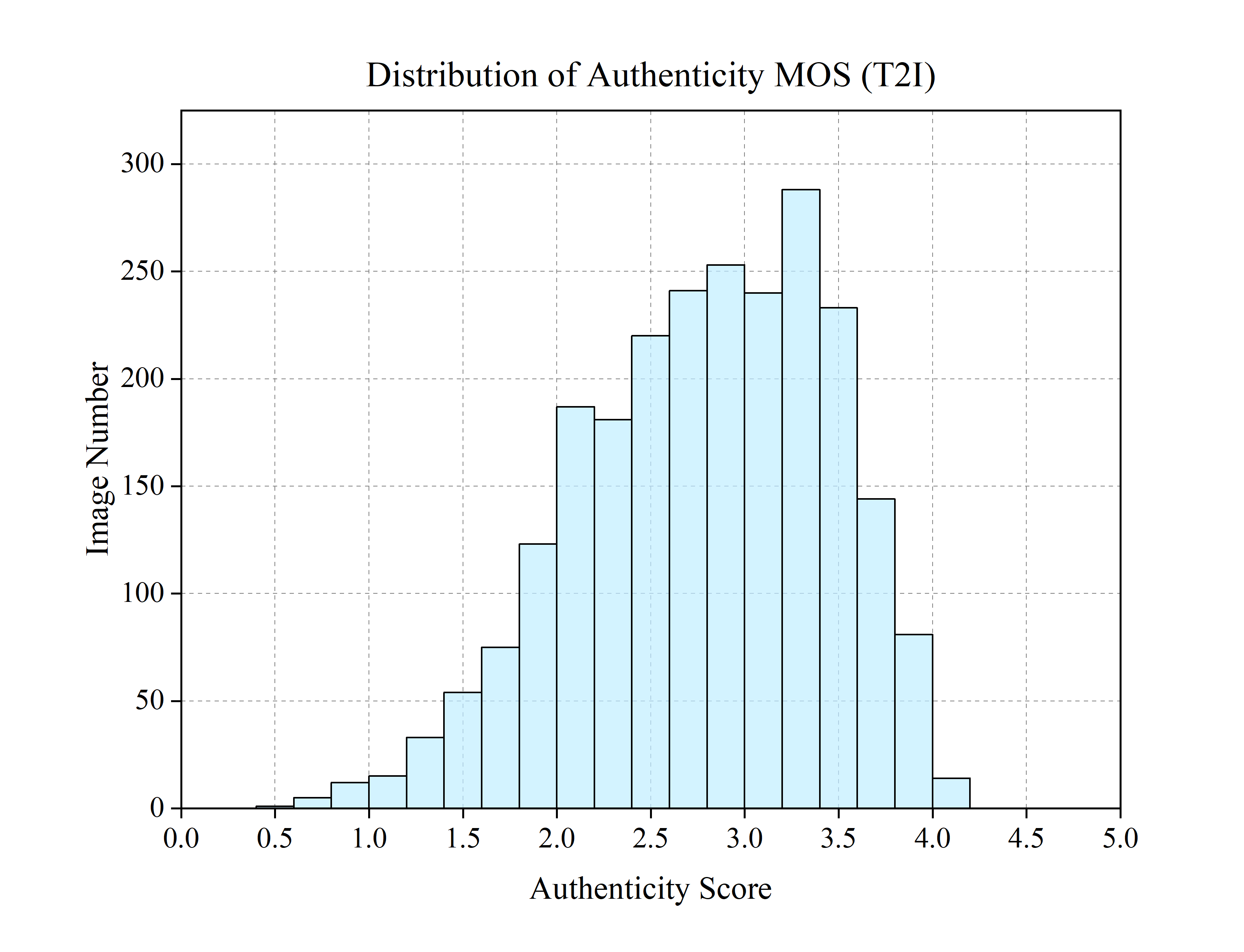}%
\label{M8}}
\hfil
\subfloat[]{\includegraphics[width=2in]{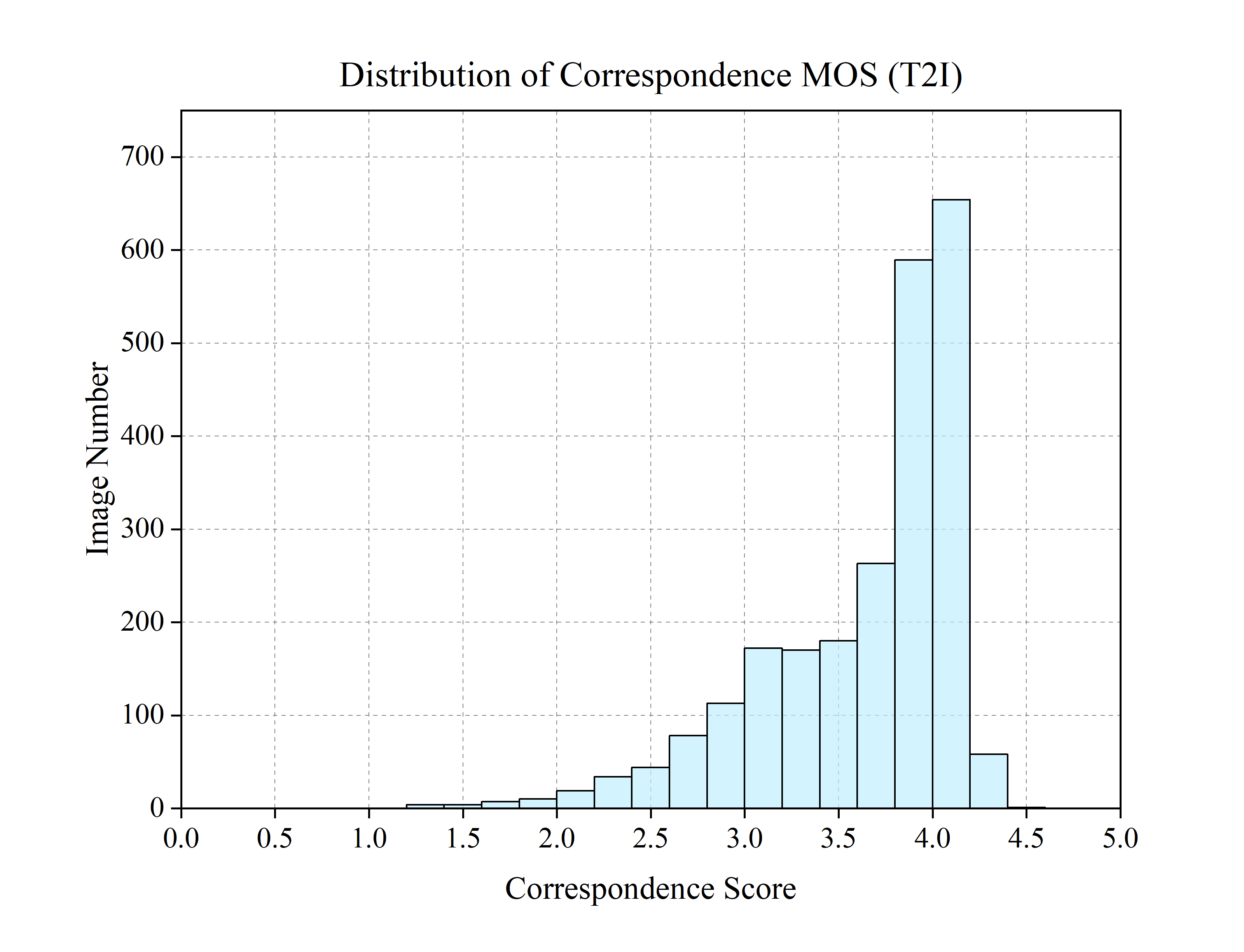}%
\label{M9}}
\hfil
\subfloat[]{\includegraphics[width=2in]{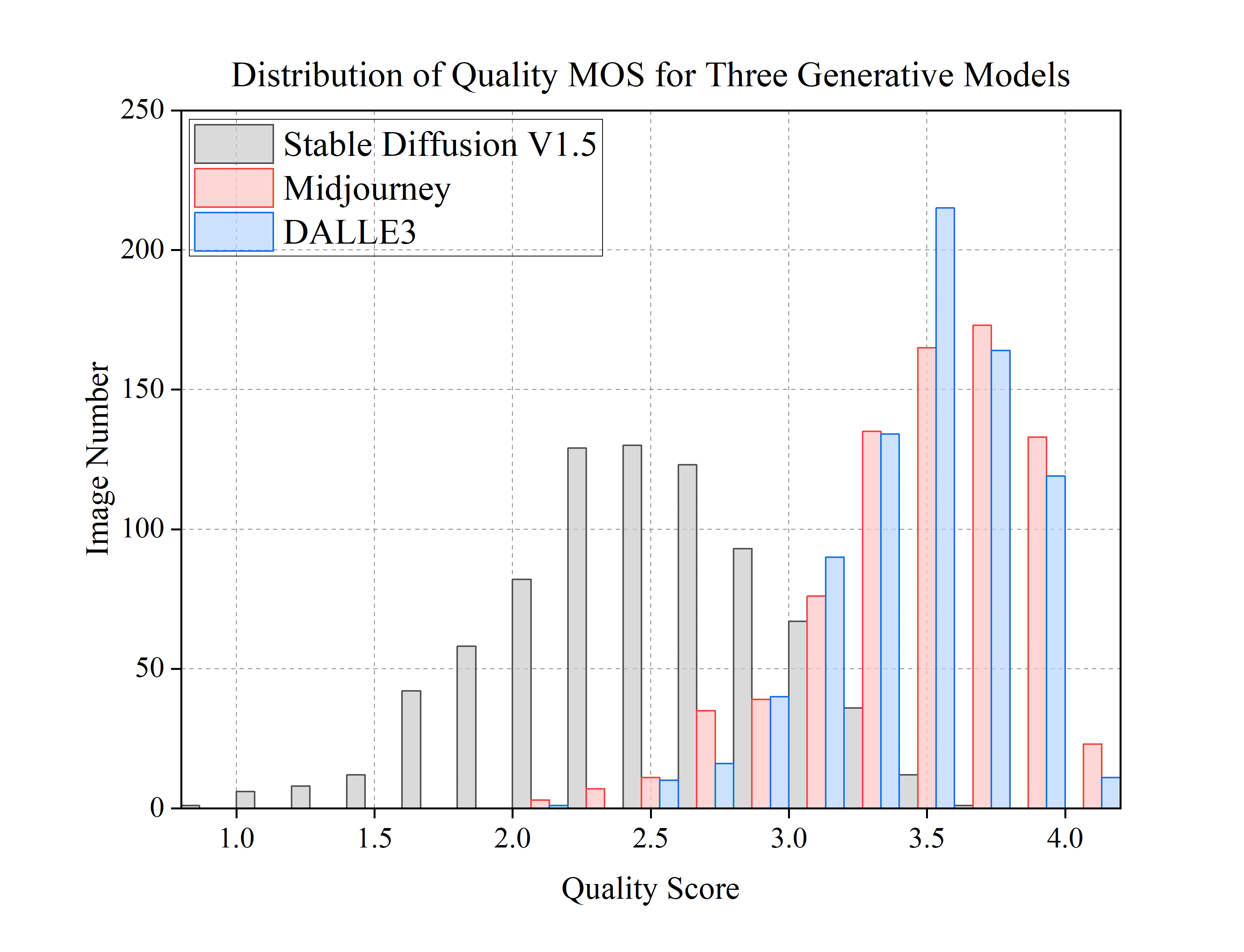}%
\label{3M1}}
\hfil
\subfloat[]{\includegraphics[width=2in]{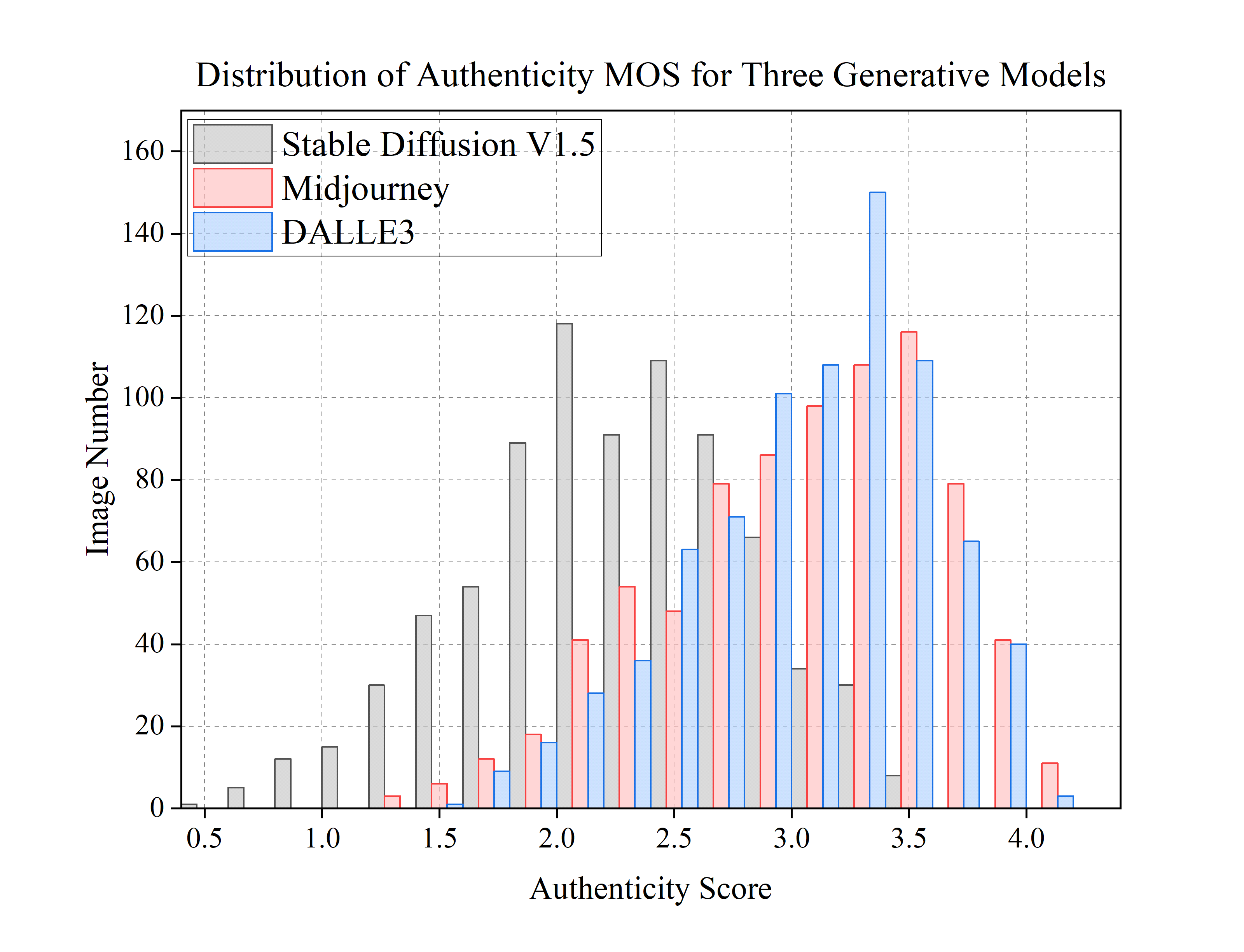}%
\label{3M2}}
\hfil
\subfloat[]{\includegraphics[width=2in]{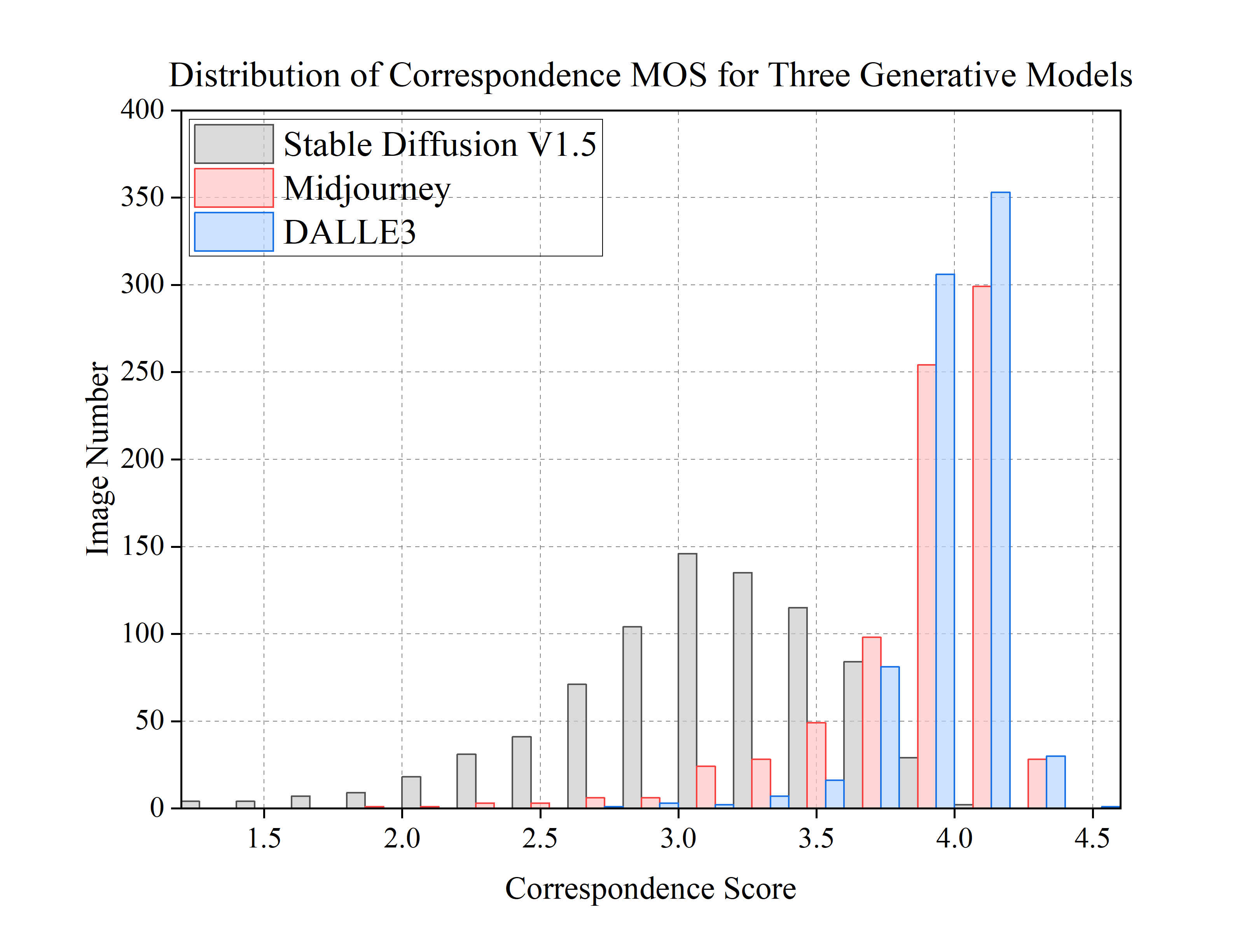}%
\label{3M3}}
\caption{Illustration of the MOS score distribution. (a), (b), and (c) exhibits the MOS distribution of quality, authenticity, and correspondence for all AIGIs in the AIGIQA-4K database, respectively.  (e), (d), and (f) exhibits the MOS distribution of quality, authenticity, and correspondence for the image-to-image AIGIs in the AIGIQA-4K database, respectively. (g), (h), and (i) exhibits the MOS distribution of quality, authenticity, and correspondence for the text-to-image AIGIs in the AIGIQA-4K database, respectively. (j), (k), and (l) exhibits the MOS distribution of quality, authenticity, and correspondence for three generative models in the AIGIQA-4K database, respectively.}
\label{M}
\end{figure*}

%\vspace{-2.5ex}
\subsection{Data Processing}
After the subjective experiments, we collect ratings from all evaluators who participate in this experiment. Following the guidelines of ITU-R BT.500-14~\cite{ITUR}, we calculate the mean and standard deviation of the subjective ratings for the same image within the same test group using the following formula:
\begin{align}
    \mu_j=\frac{1}{N}\sum_{i=1}^N r_{ij} \label{Eq1}  \\
    S_j=\sqrt{\sum_{i=1}^N \frac{(\mu_j-r_{ij})^2}{N-1}} \label{Eq2} %% 
\end{align}

The notation $r_{ij}$ represents the score of the $i_{th}$ observer for the $j_{th}$ AIGI, where $N$ denotes the total number of evaluators. When presenting the test results, all average scores should be accompanied by a relevant confidence interval, which derives from the standard deviation and the sample size. As recommended by ITU-R BT.500-14~\cite{ITUR}, we employ a $95\%$ confidence interval $( \mu_j + \epsilon_j, \mu_j - \epsilon_j )$, where $\epsilon_j$ is computed using the following formula:
\begin{align}
    \epsilon_j=1.96\cdot\frac{S_j}{\sqrt{N}} \label{Eq3} 
\end{align}
 
Scores outside the confidence interval will be considered out-of-bounds, and we will discard these scores. The mean opinion score(MOS) for the $j_{th}$ AIGI is calculated by the following formula:
\begin{align}
    MOS_j = \frac{1}{M} \sum_{i=1}^{M} r_{ij}^\prime \label{Eq4} 
\end{align}

Here, $M$ represents the number of non-discarded scores, and $r_{ij}^\prime(i=1,\ldots,M)$ denote the non-discarded scores for the $j_{th}$ AIGI.

\subsection{Database Analysis}
To further demonstrate the evaluation of AIGIs from the perspectives of quality, authenticity, and text-image correspondence, we present examples  as shown in Fig.\ref{example}. Fig.\ref{M} displays MOS score distribution for quality, authenticity, text-image correspondence. We can find that all the score distributions tend to be Gaussian distributions. Fig.\ref{M1}, Fig.\ref{M2}, and Fig.\ref{M3} demonstrate the MOS score distribution of quality, authenticity, and text-image correspondence for all AIGIs in the AIGIQA-4K database , respectively, which indicates that the database we have constructed encompasses a wide range of perceptual scores. Moreover, as shown in Fig.\ref{3M1}, Fig.\ref{3M2}, and Fig.\ref{3M3}, we can find that the images generated by Midjourney and DALLE3 are comparable in terms of quality, authenticity, and text-image correspondence, all of which are significantly superior to those generated by Stable Diffusion V1.5. Additionally, we observe that these generative models exhibit strengths and weaknesses in creating certain types of images. For example, Midjourney excels at producing high-quality images of dogs and birds, yet falls short in generating equally high-quality images of crayfish and snakes. This phenomenon could be attributed to factors such as the prompts used and the hyperparameters set. Moreover, we note distinct raw image styles inherent to the generative models employed.

\begin{figure*}[!t]
\centering
\includegraphics[width=7in]{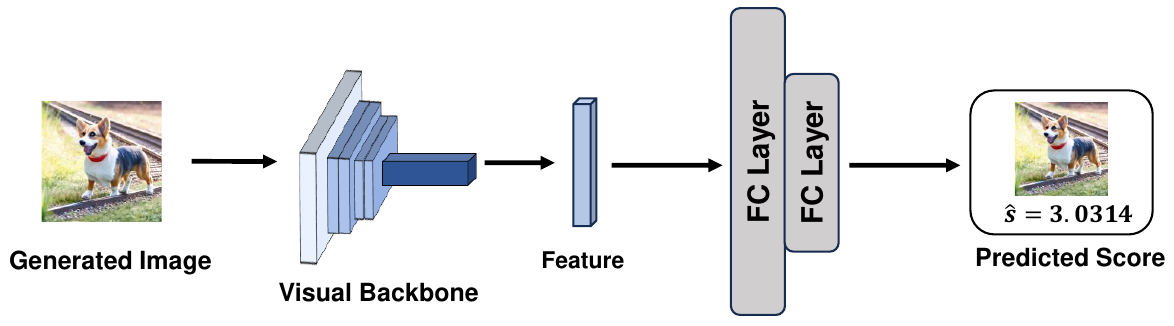}
\caption{The pipeline of the proposed NR-AIGCIQA method.}
\label{NR}
\end{figure*}

\section{Approach}
\begin{figure*}[!t]
\centering
\includegraphics[width=7in]{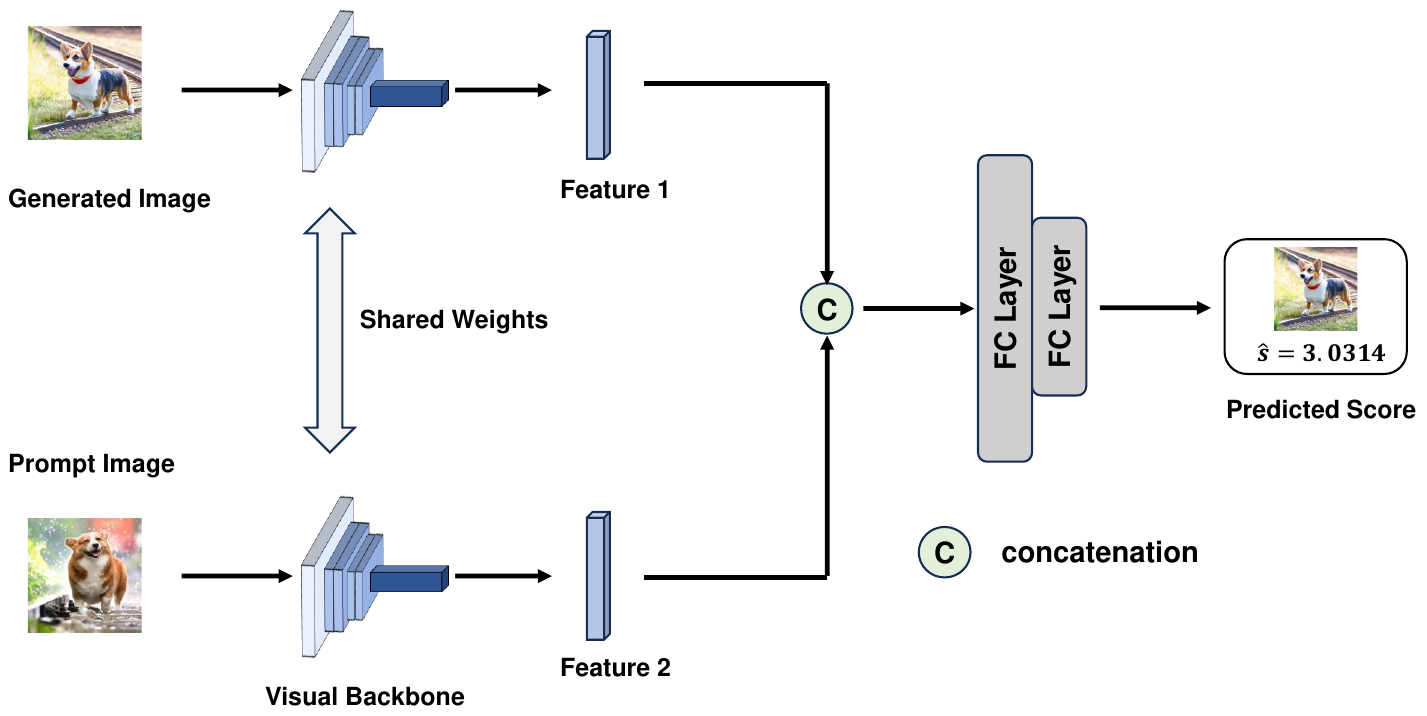}
\caption{The pipeline of the proposed FR-AIGCIQA method. .}
\label{FR}
\end{figure*}

\begin{figure*}[!t]
\centering
\includegraphics[width=7in]{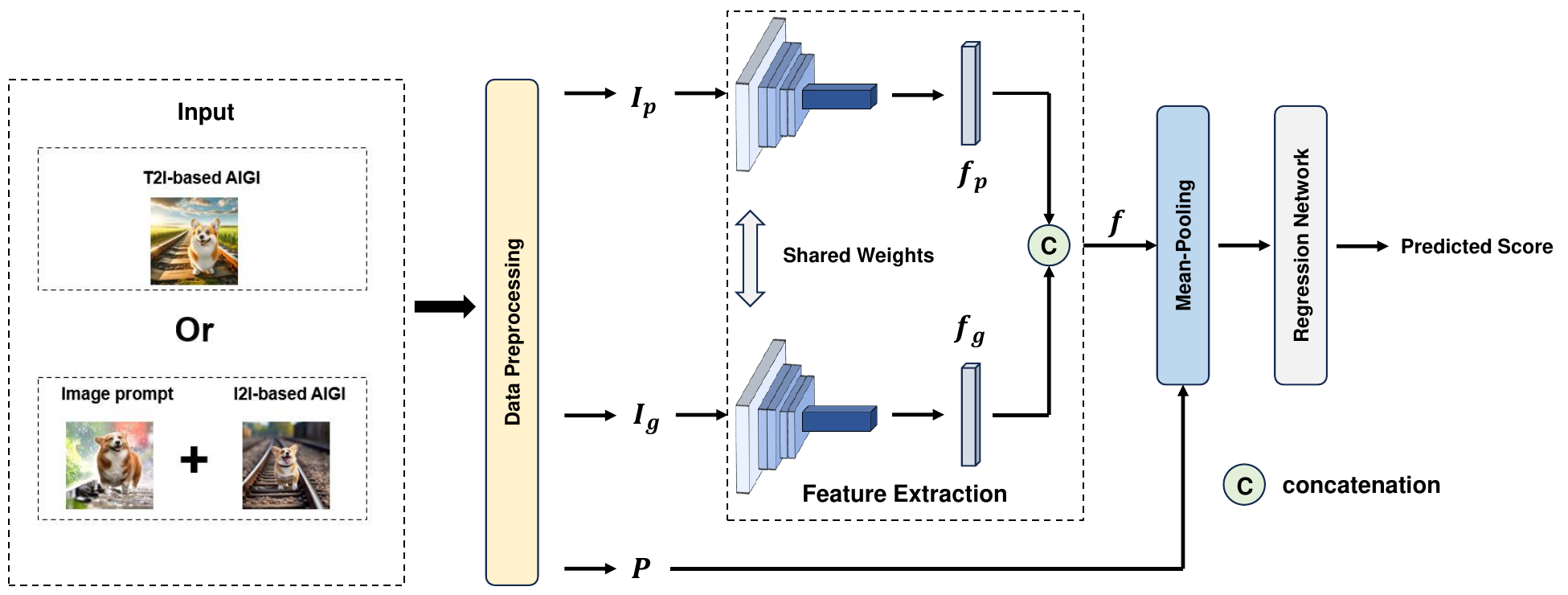}
\caption{The pipeline of the proposed PR-AIGCIQA method. .}
\label{PR}
\end{figure*}

In this section, we present three image quality assessment (IQA) methods based on pre-trained models that include a no-reference method NR-AIGCIQA, a full-reference method FR-AIGCIQA, and a partial-reference method PR-AIGCIQA. Fig.\ref{NR}, Fig.\ref{FR}, and Fig.\ref{PR} illustrate the pipelines for NR-AIGCIQA, FR-AIGCIQA, and PR-AIGCIQA methods, respectively.

\subsection{Pipeline Overview}
Given a single AIGI or a pair of AIGI and its corresponding image prompt, we first preprocess the input data. Then, we use a pre-trained visual backbone network to extract features from the input images. Finally, a regression network composed of two fully connected layers is employed to regress the predicted score. When extracting features from both the AIGIs and their corresponding image prompts simultaneously, a straightforward approach is to use a backbone network with shared weights. Our proposed benchmark methods consist of two components: a feature extraction network and a score regression network, which will be detailed as below.

\subsubsection{Feature Extraction Network}
Initially, classical image quality assessment models rely on handcrafted feature-based methods. However, the advent of convolutional neural networks has led to the predominance of deep learning-based feature extraction, which surpasses traditional methods in performance. Deep learning approaches, unlike their handcrafted counterparts that rely on empirical rules, are data-driven and excel in extracting abstract and high-level semantic features from images. In our proposed NR-AIGCIQA method and FR-AIGCIQA method, we employ several backbone network models (VGG16~\cite{simonyan2014vgg}, VGG19~\cite{simonyan2014vgg}, ResNet18~\cite{he2016resnet}, ResNet50~\cite{he2016resnet}, InceptionV4~\cite{szegedy2017inception}, and ViT\_large\_patch16\_224~\cite{dosovitskiy2020vit}) pre-trained on the ImageNet~\cite{russakovsky2015imagenet} for feature extraction from input images.

\subsubsection{Score Regression Network}
For the image features extracted by the backbone network with a feature dimension of (B, D), we employ a score regression network composed of two fully connected layers with dimensions $D \times \frac{D}{2}$  and  $\frac{D}{2} \times 1$ to regress the predictd score $\hat{s}$.

\subsubsection{Loss Function}
We optimize the parameters of the feature extraction network and the score regression network by minimizing the mean squared error between the predicted score $\hat{s}$ and the ground-truth score $s$:
\begin{align}
   L(\theta, w | I) = {MSE}(\hat{s}, s) \label{Eq5} 
\end{align}
 
Here, the parameters $\theta$ and $w$ correspond to the parameters of the regression network and the feature extraction network, respectively.

\subsection{NR-AIGCIQA}
No-reference image quality assessment (NR-IQA) methods constitute an important branch in the field of IQA task, aiming to assess the quality of images without relying on any reference images. These methods are particularly suitable for situations where reference images are unavailable or non-existent, such as when evaluating text-to-image AIGIs in this paper.

In this section, we introduce a NR-IQA method based on pre-trained models, named NR-AIGCIQA. Formally, for a given AIGI $I_g$ with score label $s$ , our proposed NR-AIGCIQA method first utilizes a visual backbone to extract features from the generated image. Subsequently, a regression network composed of two fully connected layers is employed to regress the predicted score. This method can be represented as:
\begin{align}
   \hat{s} = R_\theta(F_w(I_g)) \label{Eq6} 
\end{align}

Here, $R_\theta$ and $F_w$ denote the regression network with parameters $\theta$ and the feature extraction network with parameters $w$,respectively.

\subsection{FR-AIGCIQA}
Full-reference image quality assessment (FR-IQA) methods are widely used in the field of IQA, characterized by the need for a reference image during the assessment process. The core idea of these methods is to directly compare the differences between the image being evaluated and the reference image, thereby quantifying the quality of the image under evaluation. This approach is applicable in scenarios where a reference image is available, for example, when evaluating image-to-image AIGIs in this paper, as we can choose the image prompts as reference images.

In this section, we introduce a FR-IQA method based on pre-trained models, named FR-AIGCIQA. Formally, for a given AIGI $I_g$ with score label $s$ and the corresponding image prompt $I_p$ , our proposed FR-AIGCIQA method first employs a shared-weights backbone network to extract features from $I_g$ and $I_p$, separately. These features are then fused using concatenation, and finally, a regression network composed of two fully connected layers is applied to regress the predicted score. This method can be represented as:
\begin{align}
   \hat{s} = R_\theta(\text{Concat}(F_w(I_g), F_w(I_p))) \label{Eq7} 
\end{align}

Here, $R_\theta$ and $F_w$ denote the regression network with parameters $\theta$ and the feature extraction network with parameters $w$, respectively.

\subsection{PR-AIGCIQA}
When evaluating both text-to-image and image-to-image AIGIs, existing IQA methods often fail to fully utilize the information from image prompts. In this section, we introduce a partial-reference IQA (PR-IQA) method, named PR-AIGCIQA, which can be utilized when only a subset of images in the database has reference images available. Unlike reduced-reference IQA (RR-IQA) methods, which require only partial information or features of the reference images but necessitate that all images have reference images, the PR-IQA method proposed in this paper requires only a subset of images to have reference images, while extracting the complete image features. We will provide detailed descriptions of this method below.

Given a text-to-image AIGI or a pair of image-to-image AIGI and its corresponding image prompt, due to the inconsistency in the dimensions of input vectors, we have adopted a solution from natural language processing (NLP) for handling variable-length text inputs. Specifically, we utilize padding by appending a zero-vector of the same size to the text-to-image AIGIs, ensuring uniform dimensions across all input vectors. Since the zero-vector previously padded do not serve an actual function, we employ a padding mask vector $P$ to disregard it in subsequent computations. $P$ is defined as follows:

\begin{align}
   P = \left\{
\begin{array}{ll}
(1,1), & \text{if not padding} \\
(1,0), & \text{if padding}
\end{array}
\right.
   \label{Eq8}
\end{align}

The input vector after padding can be represented as $I=(I_g,I_p)$. Then we employ a shared-weights backbone network to extract features from $I_g$ and $I_p$. It can be expressed as:

\begin{align}
   f_g = F_w(I_g)   \label{Eq9} \\
   f_p = F_w(I_p)   \label{Eq10} 
\end{align}

Here, $f_g$ and $f_p$ denote the extracted features from $I_g$ and $I_p$, $F_w$ represents the shared-weights backbone network with parameters $w$. We use concatenation to fuse the features and obtain the fused features $f_1=(f_g,f_p)$. The fused features and padding mask vector $P=(p_0,p_1)$ are then passed to the Mean-Pooling module to obtain the final feature representation $f$. It can be expressed as:
\begin{align}
   f = \frac{f_g \times p_0 + f_p \times p_1} {p_0 + p_1} \label{Eq.4} 
\end{align}

After that, a regression network composed of two fully connected layers is employed to regress the predicted score. It can be represented as:
\begin{align}
   \hat{s} = R_\theta(f) \label{Eq11} 
\end{align}

\begin{table*}[t]
\caption{Performance comparisons of the proposed methods mentioned above and the current IQA methods on the PKU-AIGIQA-4K database. The best performance results are marked in \textcolor{red}{RED} and the second-best performance results are marked in \textcolor{blue}{BLUE}.}
\centering
%\begin{tabularx}{\textwidth}
\begin{tabular}{@{}l|cc|cc|cc@{}}
\toprule
 \multirow{2}{*}{Method} & \multicolumn{2}{c|}{Quality} & \multicolumn{2}{c|}{Authenticity} & \multicolumn{2}{c}{Correspondence} \\ 
 \cline{2-7}
 & SRCC  & PLCC  & SRCC  & PLCC & SRCC  & PLCC        \\ 
 \hline
WaDIQaM\_NR\cite{nr-fr} & 0.5693 & 0.5759 & 0.4796 & 0.4813 & 0.5277 & 0.5048 \\
CNNIQA\cite{cnriqa} & 0.4614 & 0.4354 & 0.3624 & 0.3622 & 0.3685 & 0.3578 \\
StairIQA\cite{bmsb}      & 0.7433 & 0.7404 & 0.7047 & 0.7136 & 0.7367 & 0.7261 \\
DBCNN\cite{DBCNN} & 0.6559 & 0.6504 & 0.4652 & 0.4513 & 0.5741 & 0.5497 \\
HyperIQA\cite{hyperiqa} & 0.7334 & 0.7359 & 0.6919 & 0.7041 & 0.7274 & 0.7219 \\
LinearityIQA\cite{li2020LinearityIQA} & 0.4898 & 0.4933 & 0.4153 & 0.3431 & 0.4475 & 0.2308 \\
\hline
VGG16(NR)     & 0.7411 & 0.7439 & 0.6910 & 0.6919 & 0.7277 & 0.7222 \\
VGG19(NR)     & 0.7401 & 0.7396 & 0.6808 & 0.6905 & 0.7279 & 0.7256 \\
ResNet18(NR)  & 0.7480 & 0.7394 & 0.7072 & 0.7110 & 0.7573 & 0.7432 \\
ResNet50(NR)  & 0.7424 & 0.7384 & 0.7149 & 0.7162 & 0.7329 & 0.7258 \\
InceptionV4(NR) & 0.7426 & 0.7509 & 0.7126 & 0.7157 & 0.7425 & 0.7355 \\
ViT\_large\_patch16\_224(NR)  & \textcolor{blue}{0.7646} & 0.7666 & \textcolor{red}{0.7595} & \textcolor{red}{0.7580} & \textcolor{blue}{0.7639} & \textcolor{red}{0.7919} \\
\hline
VGG16(PR)      & 0.7531 & 0.7461 & 0.7024 & 0.7044 & 0.7449 & 0.7451 \\
VGG19(PR)      & 0.7513 & 0.7520 & 0.6843 & 0.6902 & 0.7469 & 0.7495 \\
ResNet18(PR)   & 0.7582 & 0.7609 & 0.7184 & 0.7246 & 0.7611 & \textcolor{blue}{0.7828} \\
ResNet50(PR)   & 0.7640 & 0.7667 & 0.7162 & 0.7169 & 0.7442 & 0.7573 \\
InceptionV4(PR) & 0.7637 & \textcolor{red}{0.7687} & 0.7285 & 0.7317 & 0.7614 & 0.7601 \\
ViT\_large\_patch16\_224(PR)  & 0.7529 & 0.7561 & 0.7284 & 0.7305 & 0.7295 & 0.7361 \\
\hline
BERT-base+ResNet18(NR-TIER)    & 0.7369 & 0.7329 & 0.7082 & 0.7189 & 0.7514 & 0.7394 \\
BERT-base+ResNet50(NR-TIER)    & 0.7587 & 0.7504 & 0.7205 & 0.7241 & 0.7385 & 0.7437 \\
BERT-base+InceptionV4(NR-TIER) & 0.7323 & 0.7358 & 0.6965 & 0.6955 & 0.7351 & 0.7292 \\
BERT-base+ViT\_large\_patch16\_224(NR-TIER)  & 0.7623 & 0.7618 & \textcolor{blue}{0.7394} & \textcolor{blue}{0.7396} & 0.7541 & 0.7511 \\
\hline
BERT-base+ResNet18(PR-TIER)     & 0.7631 & 0.7666 & 0.7161 & 0.7229 & \textcolor{red}{0.7659} & 0.7653 \\
BERT-base+ResNet50(PR-TIER)     & 0.7624 & 0.7614 & 0.7167 & 0.7206 & 0.7498 & 0.7478 \\
BERT-base+InceptionV4(PR-TIER)  & 0.7621 & 0.7655 & 0.7163 & 0.7155 & 0.7547 & 0.7645 \\
BERT-base+ViT\_large\_patch16\_224(PR-TIER)         & \textcolor{red}{0.7650} & \textcolor{blue}{0.7673} & 0.7262 & 0.7297 & 0.6874 & 0.7066 \\
\bottomrule
\end{tabular}
\label{table2}
\end{table*}

\section{Experiment}
\subsection{Implementation Details}
Our experiments are conducted on the NVIDIA A40, using PyTorch 1.11.0 and CUDA 11.3 for both training and testing.

In the PKU-AIGIQA-4K database, scores are annotated across three dimensions: quality, authenticity, and text-image correspondence. To accurately evaluate model performance, we train individual models for each score. For feature extraction from input images, we select several backbone network models pre-trained on the ImageNet~\cite{russakovsky2015imagenet}, including VGG16~\cite{simonyan2014vgg}, VGG19~\cite{simonyan2014vgg}, ResNet18~\cite{he2016resnet}, ResNet50~\cite{he2016resnet}, InceptionV4~\cite{szegedy2017inception},  and ViT\_large\_patch16\_224~\cite{dosovitskiy2020vit}.
Due to the inconsistency in input dimensions of the backbone networks such as InceptionV4 with the image sizes in our dataset, specific preprocessing is required. For InceptionV4, we adjust image sizes to 320$\times$320, followed by random cropping to 299$\times$299 and a 50\% chance of horizontal flipping. For the other networks, images are resized to 256$\times$256, then randomly cropped to 224$\times$224 with the same probability of horizontal flipping.
During training, the batch size $B$ is set to $8$. We utilize the Adam optimizer~\cite{kingma2014adam} with a learning rate of $1 \times 10^{-4}$ and weight decay of $1 \times 10^{-5}$. The training loss employed is mean squared error (MSE) loss. In the testing phase, the batch size $B$ is set to $20$.

To evaluate the AIGI generative models in the PKU-AIGIQA-4K database, we split the data into train and test sets at a 3:1 ratio for each category produced by each generative model. We then report the performance of our  proposed methods alongside various pre-trained backbone networks and the current IQA methods. Moreover, we incorporate the TIER\cite{yuan2024tier} method to the methods presented in this paper for experiments and BERT-base\cite{Devlin2019BERT} is used to extract text features from text prompts.

We compare the performance of the following methods on the PKU-AIGIQA-4K database:

\begin{itemize}
\item \bm{$\text{The Current IQA Methods}$}: In this paper, we employ two FR-IQA methods and six NR-IQA methods for benchmark experiments. The FR-IQA methods include WaDIQaM\_FR\cite{nr-fr} and IQT\cite{cheon2021IQT}, while the NR-IQA methods include WaDIQaM\_NR\cite{nr-fr}, CNNIQA\cite{cnriqa}, StairIQA\cite{bmsb}, DBCNN\cite{DBCNN}, HyperIQA\cite{hyperiqa}, and LinearityIQA\cite{li2020LinearityIQA}.
\item \bm{$F\text{(NR)}$}: Corresponds to the NR-AIGCIQA method proposed in this paper. 
\item \bm{$F\text{(FR)}$}: Corresponds to the FR-AIGCIQA method proposed in this paper. 
\item \bm{$F\text{(PR)}$}: Corresponds to the PR-AIGCIQA method proposed in this paper. 
\item \bm{$\text{BERT-base}+F\text{(NR-TIER)}$}: Incorporating the TIER method to the NR-AIGCIQA method. 
\item \bm{$\text{BERT-base}+F\text{(FR-TIER)}$}: Incorporating the TIER method to the FR-AIGCIQA method. 
\item \bm{$\text{BERT-base}+F\text{(PR-TIER)}$}: Incorporating the TIER method to the PR-AIGCIQA method.

\end{itemize}

Here, $F$ denotes the visual backbone.

\begin{table*}[!h]
\caption{Performance comparisons of the proposed methods mentioned above and the current IQA methods on the T2IQA subset of PKU-AIGIQA-4K database. The best performance results are marked in \textcolor{red}{RED} and the second-best performance results are marked in \textcolor{blue}{BLUE}.}
\centering
%\begin{tabularx}{\textwidth}
\begin{tabular}{@{}l|cc|cc|cc@{}}
\toprule
 \multirow{2}{*}{Method} & \multicolumn{2}{c|}{Quality} & \multicolumn{2}{c|}{Authenticity} & \multicolumn{2}{c}{Correspondence} \\ 
 \cline{2-7}
 & SRCC  & PLCC  & SRCC  & PLCC & SRCC  & PLCC        \\ 
\hline
WaDIQaM\_NR\cite{nr-fr}   & 0.6799 & 0.6745 & 0.5506 & 0.5242 & 0.6264 & 0.6618 \\
CNNIQA\cite{cnriqa}        & 0.5924 & 0.5921 & 0.4980 & 0.4894 & 0.5450 & 0.5408 \\
StairIQA\cite{bmsb}     & 0.8196 & 0.8121 & 0.7715 & 0.7625 & 0.7858 & 0.7955 \\
DBCNN\cite{DBCNN}         & 0.7251 & 0.7385 & 0.6774 & 0.6639 & 0.6757 & 0.7251 \\
HyperIQA\cite{hyperiqa}      & 0.8146 & 0.8091 & 0.7724 & 0.7666 & 0.7859 & 0.7962 \\
LinearityIQA\cite{li2020LinearityIQA}  & 0.6109 & 0.5226 & 0.5369 & 0.2527 & 0.5974 & 0.5649 \\
\hline
VGG16(NR)       & 0.8199 & 0.8119 & 0.7529 & 0.7413 & 0.7798 & 0.7872 \\
VGG19(NR)      & 0.8174 & 0.8084 & 0.7498 & 0.7423 & 0.7839 & 0.7865 \\
ResNet18(NR)    & 0.8128 & 0.8018 & 0.7815 & 0.7710 & 0.7879 & 0.7934 \\
ResNet50(NR)    & 0.8138 & 0.8086 & 0.7717 & 0.7618 & 0.7811 & 0.7837 \\
InceptionV4(NR) & 0.8099 & 0.8067 & 0.7643 & 0.7419 & 0.7725 & 0.7746 \\
ViT\_large\_patch16\_224(NR)    & 0.8299 & \textcolor{blue}{0.8329} & \textcolor{blue}{0.7913} & \textcolor{blue}{0.7821} & \textcolor{blue}{0.7907} & \textcolor{red}{0.8249} \\
\hline
BERT-base+ResNet18(NR-TIER)    & 0.8136 & 0.8102 & 0.7817 & 0.7682 & \textcolor{red}{0.7940} & 0.8007 \\
BERT-base+ResNet50(NR-TIER)    & 0.8197 & 0.8178 & 0.7792 & 0.7686 & 0.7819 & 0.7865 \\
BERT-base+InceptionV4(NR-TIER) & \textcolor{blue}{0.8317} & 0.8314 & 0.7783 & 0.7653 & 0.7901 & 0.8077 \\
BERT-base+ViT\_large\_patch16\_224(NR-TIER)    & \textcolor{red}{0.8356} & \textcolor{red}{0.8383} & \textcolor{red}{0.7997} & \textcolor{red}{0.7921} & 0.7680 & \textcolor{blue}{0.8110} \\
\bottomrule
\end{tabular}
\label{table3}
\end{table*}

\begin{table*}[!h]
\caption{Performance comparisons of the proposed methods mentioned above and the current IQA methods on the I2IQA subset of PKU-AIGIQA-4K database. The best performance results are marked in \textcolor{red}{RED} and the second-best performance results are marked in \textcolor{blue}{BLUE}.}
\centering
%\begin{tabularx}{\textwidth}
\begin{tabular}{@{}l|cc|cc|cc@{}}
\toprule
 \multirow{2}{*}{Method} & \multicolumn{2}{c|}{Quality} & \multicolumn{2}{c|}{Authenticity} & \multicolumn{2}{c}{Correspondence} \\ 
 \cline{2-7}
 & SRCC  & PLCC  & SRCC  & PLCC & SRCC  & PLCC        \\ 
 \hline
WaDIQaM\_NR\cite{nr-fr} & 0.4261 & 0.4269 & 0.3983 & 0.4106 & 0.4425 & 0.4425 \\
CNNIQA\cite{cnriqa} & 0.2180 & 0.2412 & 0.3110 & 0.3511 & 0.2551 & 0.2893 \\
StairIQA\cite{bmsb}                & 0.6819     & 0.6999    & 0.6362      & 0.7035     & 0.7483        & 0.7438         \\
DBCNN\cite{DBCNN} & 0.4891 & 0.5207 & 0.4619 & 0.5126 & 0.4276 & 0.3997 \\
HyperIQA\cite{hyperiqa} & 0.7020 & 0.7204 & 0.6532 & 0.7037 & 0.7100 & 0.6847 \\
LinearityIQA\cite{li2020LinearityIQA} & 0.3638 & 0.2363 & 0.3629 & 0.2218 & 0.4711 & 0.4881 \\
\hline
WaDIQaM\_FR\cite{nr-fr} & 0.5096 & 0.5165 & 0.4915 & 0.5185 & 0.6146 & 0.6400 \\
IQT\cite{cheon2021IQT} & 0.6584 & 0.6589 & 0.6142 & 0.6625 & 0.6762 & 0.7251 \\
\hline
VGG16(NR)        & 0.6734     & 0.6854    & 0.6449      & 0.6975     & 0.7130        & 0.7095         \\
VGG19(NR)                & 0.6836     & 0.6855    & 0.6352      & 0.6845     & 0.7383        & 0.7348        \\
ResNet18(NR)          & 0.6885     & 0.7112    & 0.6684      & 0.7108     & 0.7492        & 0.7317         \\
ResNet50(NR)               & 0.6876     & 0.6875    & 0.6530      & 0.6918     & 0.7456        & 0.7385         \\
InceptionV4(NR)            & 0.6988     & 0.7076    & 0.6733      & 0.7191     & 0.7509        & 0.7306        \\ 
ViT\_large\_patch16\_224(NR)   & 0.7217     & 0.7256    & 0.7059      & \textcolor{blue}{0.7484}     & 0.7982        & 0.8014        \\
\hline
VGG16(FR)                   & 0.6825     & 0.6918    & 0.6468      & 0.7005     & 0.7589        & 0.7740        \\
VGG19(FR)                   & 0.6832     & 0.7093    & 0.6505      & 0.7056     & 0.7594        & 0.7741         \\
ResNet18(FR)                & 0.7063    & 0.7249    & 0.6724      & 0.7220     & 0.7737       & 0.7892          \\
ResNet50(FR)               & 0.6885     & 0.6968    & 0.6567      & 0.6983     & 0.7662        & 0.7803         \\
InceptionV4(FR)    & 0.7017     & 0.7246    & 0.6788      & 0.7298     & 0.7626        & 0.7627   \\
ViT\_large\_patch16\_224(FR)      & \textcolor{red}{0.7467}    & 0.7337  & \textcolor{red}{0.7287}     & \textcolor{red}{0.7623}     & 0.8023       & \textcolor{blue}{0.8148}      \\
\hline
BERT-base+ResNet18(NR-TIER)               & 0.6788    & 0.7104    &    0.6518   &   0.6968   &      0.7453   &   0.7399      \\
BERT-base+ResNet50(NR-TIER)              &  0.6953  &  0.7003   &  0.6543    &  0.7055    &     0.7434    &   0.7360    \\
BERT-base+InceptionV4(NR-TIER)    &   0.6862   &   0.7187 &    0.6274   &   0.6842   &    0.7440    & 0.7382      \\
BERT-base+ViT\_large\_patch16\_224(NR-TIER)               & 0.7275     & \textcolor{blue}{0.7363}    & 0.7089      & 0.7447     & \textcolor{blue}{0.8123}        & 0.8051       \\
\hline
BERT-base+ResNet18(FR-TIER)               & 0.7082    & 0.7097    & 0.6681      & 0.7155     & 0.7825        & 0.7938         \\
BERT-base+ResNet50(FR-TIER)               & 0.6869     & 0.7031    & 0.6710      & 0.7106     & 0.7752        & 0.7894         \\
BERT-base+InceptionV4(FR-TIER)    & 0.6997     & 0.7245    & 0.6675      & 0.7081     & 0.7725        & 0.7840     \\
BERT-base+ViT\_large\_patch16\_224(FR-TIER)              & \textcolor{blue}{0.7446}     & \textcolor{red}{0.7430}    & \textcolor{blue}{0.7145}      & 0.7466     & \textcolor{red}{0.8149}        & \textcolor{red}{0.8266}     \\
\bottomrule
\end{tabular}
\label{table4}
\end{table*}

\subsection{Evaluation Criteria}
Following prior research, we utilize the Spearman rank correlation coefficient (SRCC) and Pearson linear correlation coefficient (PLCC) as evaluation metrics to evaluate the performance of our model.

The SRCC is defined as follows:
\begin{align}
   \text{SRCC} = 1 - \frac{6 \sum_{i=1}^{N} d_i^2}{N(N^2 - 1)} \label{Eq12} 
\end{align}

Here, $N$ represents the number of test images, and $d_i$ denotes the difference in ranking between the true quality scores and the predicted quality scores for the $i_{th}$ test image.

The PLCC is defined as follows:
\begin{align}
    \text{PLCC} = \frac{\sum_{i=1}^{N}(s_i - \mu_{s_i})(\hat{s}_i - \hat{\mu}_{s_i})}{\sqrt{\sum_{i=1}^{N}(s_i - \mu_{s_i})^2 \sum_{i=1}^{N}(\hat{s}_i - \hat{\mu}_{s_i})^2}}\label{Eq13} 
\end{align}

Here, $s_i$ and $\hat{s}_i$ represent the true and predicted quality scores for the $i_{th}$ image, respectively. $\mu_{s_i}$ and $\hat{\mu}_{s_i}$ are their respective means, and $N$ is the number of test images.
Both SRCC and PLCC are metrics used to evaluate the relationship between two sets of variables. They range between $-1$ and $1$, where a positive value indicates a positive correlation and a negative value indicates a negative correlation, and a larger value means a better performance.

\subsection{Results and Analysis}
 The performance results of the proposed methods and the current IQA methods on the PKU-AIGIQA-4K database are exhibited in TABLE \ref{table2}, TABLE \ref{table3}, and TABLE \ref{table4}. TABLE \ref{table2} shows the performance results on all images of the PKU-AIGIQA-4K database, TABLE \ref{table3} shows the performance results on T2IQA subset of the PKU-AIGIQA-4K database, and TABLE \ref{table4} shows the performance results on I2IQA subset of the PKU-AIGIQA-4K database. 
 Based on the results reported, we can draw several conclusions:

\begin{itemize}

\item Among the backbone networks we utilize, ViT\_large\_patch16\_224~\cite{dosovitskiy2020vit} exhibits the best performance, significantly surpassing other backbone networks.

\item Based on the results reported in TABLE \ref{table2}, We can find that the benchmark model of the PR-AIGCIQA method outperforms the benchmark model of NR-AIGCIQA method in most cases. This demonstrates that our proposed PR-AIGCIQA method can effectively enhance the performance of the model by fully utilizing the information from image prompts. However, when employing ViT\_large\_patch16\_224 as the visual backbone network for feature extraction, the performance of the PR-AIGCIQA method falls short compared to the NR-AIGCIQA method.

\item Based on the results reported in TABLE \ref{table4}, We can find that the benchmark model of the FR-AIGCIQA method outperforms the benchmark model of NR-AIGCIQA method in most cases. This demonstrates that incorporating image prompts as reference images can effectively improve the performance of the model.

\item The experimental results indicate that the proposed methods and the current IQA methods do not adequately align with human preferences for AIGIs. 

\end{itemize}

\subsection{Case Study}
\begin{figure}[!t]
\centering
\subfloat[]{\includegraphics[width=3.5in]{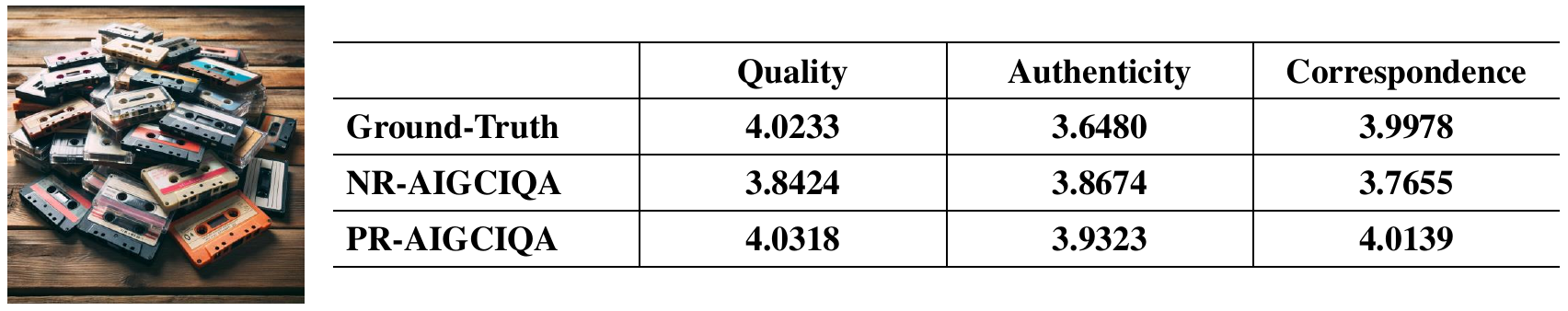}%
\label{Case1}}
\hfil
\subfloat[]{\includegraphics[width=3.5in]{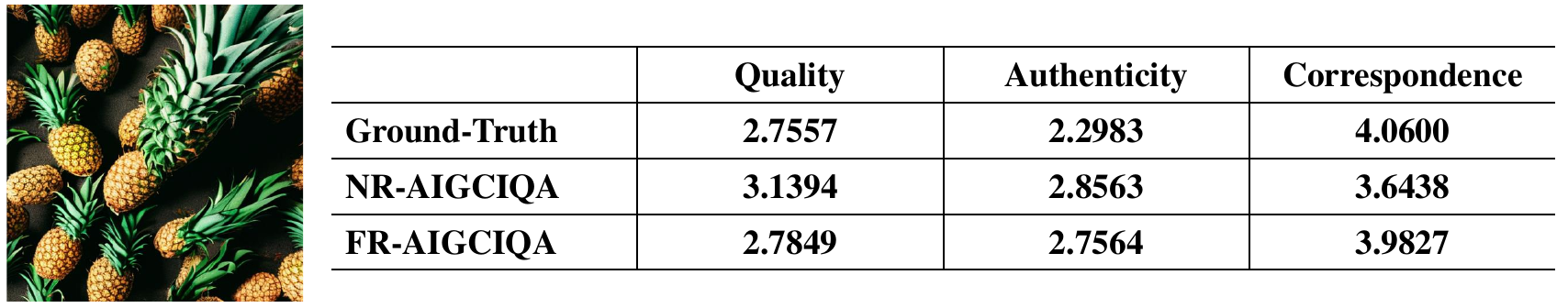}%
\label{Case2}}
\caption{Case study with qualitative results, which present the comparisons of the proposed methods and ground-truth. (a) present the comparisons of the NR-AIGCIQA, PR-AIGCIQA and ground-truth; (b) present the comparisons of the NR-AIGCIQA, FR-AIGCIQA and ground-truth.}
\label{Case}
\end{figure}

We further conduct a case study to analyze the behavior of our proposed method, as shown in Fig.\ref{Case} Case (a)  employs a text-to-image AIGI generated by DALLE3, where the final scores predicted by the PR-AIGCIQA method are closer to the ground-truth in terms of quality and correspondence, while the final scores predicted by the NR-AIGCIQA method are closer to the ground-truth in terms of authenticity. Case (b) uses an image-to-image AIGI produced by Stable Diffusion V1.5, where the final scores predicted by the FR-AIGCIQA method are closer to the ground-truth across quality, authenticity, and correspondence compared to those predicted by the NR-AIGCIQA method.

\section{Conclusion}
In recent years, image generation technology has rapidly advanced, leading to a proliferation of AIGIs of varying quality across various social media platforms. Consequently, developing efficient methods to evaluate AIGIs has become increasingly important. Previous research primarily focus on evaluating the quality of text-to-image (T2I) AIGIs. In this paper, we conduct a broader exploration of evaluating the quality of both T2I and image-to-image (I2I) AIGIs. We first establish a large-scale perceptual quality assessment database PKU-AIGIQA-4K, which includes both T2I and I2I AIGIs, totaling 4,000 images with 2,400 T2I and 1,600 I2I AIGIs. Subsequently, we conduct a well-organized subjective experiment to collect quality labels for AIGIs and perform a comprehensive analysis of the PKU-AIGIQA-4K database. Regarding the usage of image prompts during the training and testing processes, we propose a no-reference image quality assessment (NR-IQA) method NR-AIGCIQA, a full-reference image quality assessment (FR-IQA) method FR-AIGCIQA, and a novel partial-reference image quality assessment (PR-IQA) method PR-AIGCIQA based on pre-trained models. Finally, we perform benchmark experiments using the PKU-AIGIQA-4K database to evaluate the performance of the proposed methods and the current IQA methods. We hope that the PKU-AIGIQA-4K database and the proposed AIGIQA methods can provide insights and inspiration for future AIGIQA research.

\bibliographystyle{IEEEtran}
\bibliography{cite.bib}

\vfill

\end{document}